\documentclass{article}
\usepackage{arxiv}
\newcommand\alumina{$\alpha$-Al\textsubscript{2}O\textsubscript{3}}

\usepackage[utf8]{inputenc} 
\usepackage[T1]{fontenc}    
\usepackage{hyperref}       
\usepackage{url}            
\usepackage{booktabs}       
\usepackage{amsfonts}       
\usepackage{nicefrac}       
\usepackage{microtype}      
\usepackage{graphicx}
\usepackage[numbers,square]{natbib}
\usepackage{amssymb}
\usepackage{amsmath}
\usepackage{caption}
\usepackage{csquotes}
\usepackage{subcaption}
\usepackage{adjustbox}
\usepackage{underscore}
\usepackage[pagewise, mathlines]{lineno}
\usepackage[capitalise]{cleveref}
\usepackage{atbegshi}
\AtBeginDocument{\AtBeginShipoutNext{\AtBeginShipoutDiscard}}

\title{Integrating Unsupervised and Supervised learning approaches to unveil Critical Process Inputs}

\author{\href{https://orcid.org/0000-0002-3136-1402}{\includegraphics[scale=0.06]{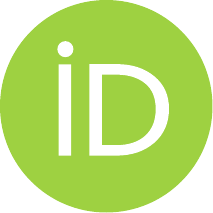}\hspace{1mm}Paris Papavasileiou\thanks{Author also affiliated with the School of Chemical Engineering, National Technical University of Athens, Zographos Campus, 15780, Attiki, Greece}}\\
	Faculty of Science, Technology and Medicine\\
	University of Luxembourg\\
	Esch-sur-Alzette, L-4364, Luxembourg\\
    \texttt{paris.papavasileiou@uni.lu}\\
	\And
	\href{https://orcid.org/0000-0003-2272-2584}{\includegraphics[scale=0.06]{orcid.pdf}\hspace{1mm}Dimitrios G.~Giovanis} \\
	Department of Civil \& Systems Engineering, Whiting School of Engineering\\
    Johns Hopkins University\\
    Baltimore, MD 21218, USA\\
	\texttt{dgiovan1@jhu.edu}\\
    \And
	Martin Kathrein \\
	CERATIZIT Luxembourg S.à r.l.\\
    Mamer, L-8201, Luxembourg\\
	\texttt{Martin.Kathrein@ceratizit.com} \\
    \And
	Gabriele Pozzetti \\
	CERATIZIT Luxembourg S.à r.l.\\
    Mamer, L-8201, Luxembourg\\
	\texttt{Gabriele.Pozzetti@plansee-group.com} \\
    \And
	Christoph Czettl \\
	CERATIZIT Austria GmbH\\
    Reutte, A-6600, Austria\\
	\texttt{Christoph.Czettl@ceratizit.com} \\ 
    \And
	\href{https://orcid.org/0000-0003-2220-3522}{\includegraphics[scale=0.06]{orcid.pdf}\hspace{1mm}Ioannis G.~Kevrekidis} \\
	Department of Chemical and Biomolecular Engineering \\ \& Department of Applied Mathematics and Statistics, Whiting School of Engineering\\
    Johns Hopkins University\\
    Baltimore, MD 21218, USA\\
	\texttt{yannisk@jhu.edu}\\
    \And
	\href{https://orcid.org/0000-0001-6651-7318}{\includegraphics[scale=0.06]{orcid.pdf}\hspace{1mm}Andreas G.~Boudouvis} \\
	School of Chemical Engineering\\
    National Technical University of Athens\\
    Zographos Campus, 15780, Attiki, Greece\\
	\texttt{boudouvi@chemeng.ntua.gr} \\
    \And
	\href{https://orcid.org/0000-0001-7622-2193}{\includegraphics[scale=0.06]{orcid.pdf}\hspace{1mm}St\'{e}phane P.A.~Bordas} \\
	Faculty of Science, Technology and Medicine\\
	University of Luxembourg\\
	Esch-sur-Alzette, L-4364 \\
	\texttt{stephane.bordas@uni.lu}  \\ 
    \And
    \href{https://orcid.org/0000-0002-5229-4157}{\includegraphics[scale=0.06]{orcid.pdf}\hspace{1mm}Eleni D.~Koronaki} \\
	Faculty of Science, Technology and Medicine\\
	University of Luxembourg\\
	Esch-sur-Alzette, L-4364, Luxembourg\\
	\texttt{eleni.koronaki@uni.lu}
}

\renewcommand{\shorttitle}{Papavasileiou et al. (2024)}

\hypersetup{
pdftitle={Integrating Unsupervised and Supervised learning approaches to unveil Critical Process Inputs},
pdfsubject={},
pdfauthor={Paris Papavasileiou, Eleni D.~Koronaki},
pdfkeywords={critical parameters, machine learning, industrial process, data-driven approaches, chemical vapor deposition}
}

\begin{document}
\maketitle
\begin{abstract}
This study introduces a machine learning framework tailored to large-scale industrial processes characterized by a plethora of numerical and categorical inputs. The framework aims to (i) discern critical parameters influencing the output and (ii) generate accurate out-of-sample qualitative and quantitative predictions of production outcomes. Specifically, we address the pivotal question of the significance of each input in shaping the process outcome, using an industrial Chemical Vapor Deposition (CVD) process as an example. The initial objective involves merging subject matter expertise and clustering techniques exclusively on the process output, here, coating thickness measurements at various positions in the reactor. This approach identifies groups of production runs that share similar qualitative characteristics, such as film mean thickness and standard deviation. In particular, the differences of the outcomes represented by the different clusters can be attributed to differences in specific inputs, indicating that these inputs are potentially critical for the production outcome. Shapley Value analysis corroborates the formed hypotheses. Leveraging this insight, we subsequently implement supervised classification and regression methods using the identified critical process inputs. The proposed methodology proves to be valuable in scenarios with a multitude of inputs and insufficient data for the direct application of deep learning techniques, providing meaningful insights into the underlying processes.
\end{abstract}


\section{Introduction}
\label{sec:intro}
Chemical vapor deposition (CVD) is a widely used chemical process for producing thin films with various properties, applied in semiconductor manufacturing \citep{cotePlasmaassistedChemicalVapor1999,biefeldMetalorganicChemicalVapor2002}, membranes \citep{haPropertiesTiO2Membranes1996,khatibSilicaMembranesHydrogen2013}, protective \citep{schmauderHardCoatingsPlasma2006,jiaCVDGrowthHighquality2021} and wear-resistant \citep{karnerCVDDiamondCoated1996,kathreinDopedCVDAl2O32003} coatings. Although Computational Fluid Dynamics (CFD) models traditionally explore CVD complexity \citep{mitrovicProcessConditionsOptimization2007,cheimariosIlluminatingNonlinearDependence2012,koronakiNonAxisymmetricFlowFields2014,gakisNumericalInvestigationMultiple2015,koronakiEfficientTracingStability2016,aviziotisMultiscaleModelingExperimental2016,aviziotisCombinedMacroNanoscale2017,psarellisInvestigationReactionMechanisms2018,topkaInnovativeKineticModel2022}
their efficiency and adequacy are challenged in cases involving unknown chemical reactions or intricate reactor geometries. The computational cost of large-scale industrial process models and the nonlinear nature of competing physical and chemical mechanisms further limit the utility of CFD as a viable \enquote{digital twin}. It is also possible that there are different process outputs arise for the same inputs, which is also linked to non-linearity \citep{gkinisEffectsFlowMultiplicity2017,koronakiClassificationStatesModel2019}.

Recently, Machine Learning (ML) has emerged as a promising alternative in the era of Industry 4.0 with abundant process data. ML applications range from maintenance management \citep{saxenaEvolvingArtificialNeural2007,sustoMachineLearningPredictive2015,wuExperimentalStudyProcess2019} and production planning \citep{prioreLearningbasedSchedulingFlexible2018,agarwalDeepLearningClassification2020} to outcome prediction \citep{papananiasBayesianFrameworkEstimate2019,papavasileiouEquationbasedDatadrivenModeling2023}, process control \citep{maContinuousControlPolymerization2019}, and optimization \citep{humfeldMachineLearningFramework2021}. ML models can also be developed based on preexisting physics-based models, in order to further investigate the modeled process \citep{gkinisBuildingDatadrivenReduced2019, spencerInvestigationChemicalVapor2021, martin-linaresPhysicsagnosticPhysicsinfusedMachine2023}.



Despite recent advances in explainable AI (XAI), challenges persist in addressing the \enquote{black box} nature of ML models. However, tools such as SHAP (SHapley Additive exPlanations) offer improved explainability using a game theory approach \citep{shapleyValueNPersonGames1952,lundbergUnifiedApproachInterpreting2017,barredoarrietaExplainableArtificialIntelligence2020a,sundararajanManyShapleyValues2020}.

This study utilizes production data from an industrial CVD reactor for wear-resistant coating production of cutting tools. The data encompass details about reactor setup and process inputs; thickness measurements of the Ti(C,N)/\alumina{} coating in 15 positions within the reactor are considered as process outputs.

Implementing state-of-the-art (SotA) methods faces challenges that include:

\begin{itemize}
    \item Process complexity, namely, multiscale interacting phenomena in intricate geometries.
    \item A multitude of numerical and categorical inputs, with little insight of their impact on the process outcome.
    \item Noisy and heterogeneous data, collected over months or years with varying instrumentation and calibration, which cannot be categorized as \enquote{big}.
\end{itemize}

Several different options are available in the literature related to the discovery of important process parameters and the facilitation of subsequent modeling attempts. Variable Importance in Projection (VIP) parameters \citep{chongPerformanceVariableSelection2005,luIndustrialPLSModel2014}, a byproduct of Partial Least Squares (PLS) models, have traditionally been used to determine the impact of process inputs on the output \citep{garthwaiteInterpretationPartialLeast1994,luIndustrialPLSModel2014,kumarPartialLeastSquare2021}. 

 Variable selection tools have been shown to enable improved performance and subsequently lead to a greater understanding of the importance of input variables on model output \citep{heinzeVariableSelectionReview2018}. To this end, several powerful dimensionality reduction techniques based on Principal Component Analysis (PCA) or Diffusion Maps (DMaps) \citep{koronakiDatadrivenReducedorderModel2020,koronakiPartialDataOutofsample2023a} can lead to the discovery of effective process parameters \citep{brouwerUnderlyingConnectionsIdentifiability2018,evangelouParameterCombinationsThat2022}.

Despite the effectiveness of existing methods for strictly numerical data, challenges arise when dealing with datasets rich in categorical features, as seen in this application. This work aims to propose an ML workflow for the identification of critical process inputs without labeled data, an essential contribution to control, optimization, and experimental design.

Our approach involves an unsupervised analysis of process outputs to identify clusters of similar production runs. Subsequently, we analyze relevant process input data to discern distinguishing characteristics within these clusters. Our findings are supported by subject matter expertise. Shifting to supervised learning, we use cluster labels to train a classifier for predicting these labels given specific process inputs. Furthermore, we attempt to create a regression model for predicting thickness measurements. Finally, we employ SHAP and Shapley values to interpret the model output.

The manuscript is structured as follows. A brief overview of the process and the available production data is given in \cref{sec:process-overview}. The various machine learning methods implemented (supervised, unsupervised) are presented in \cref{sec:ml-tools}. The results are discussed in \cref{sec:results}, followed by concluding remarks in \cref{sec:conclusion}.

\section{Process overview}
\label{sec:process-overview}
\begin{figure}[h]
\captionsetup[subfigure]{justification=centering}
\centering
\begin{subfigure}[b]{0.48\textwidth}
   \includegraphics[width=1\textwidth]{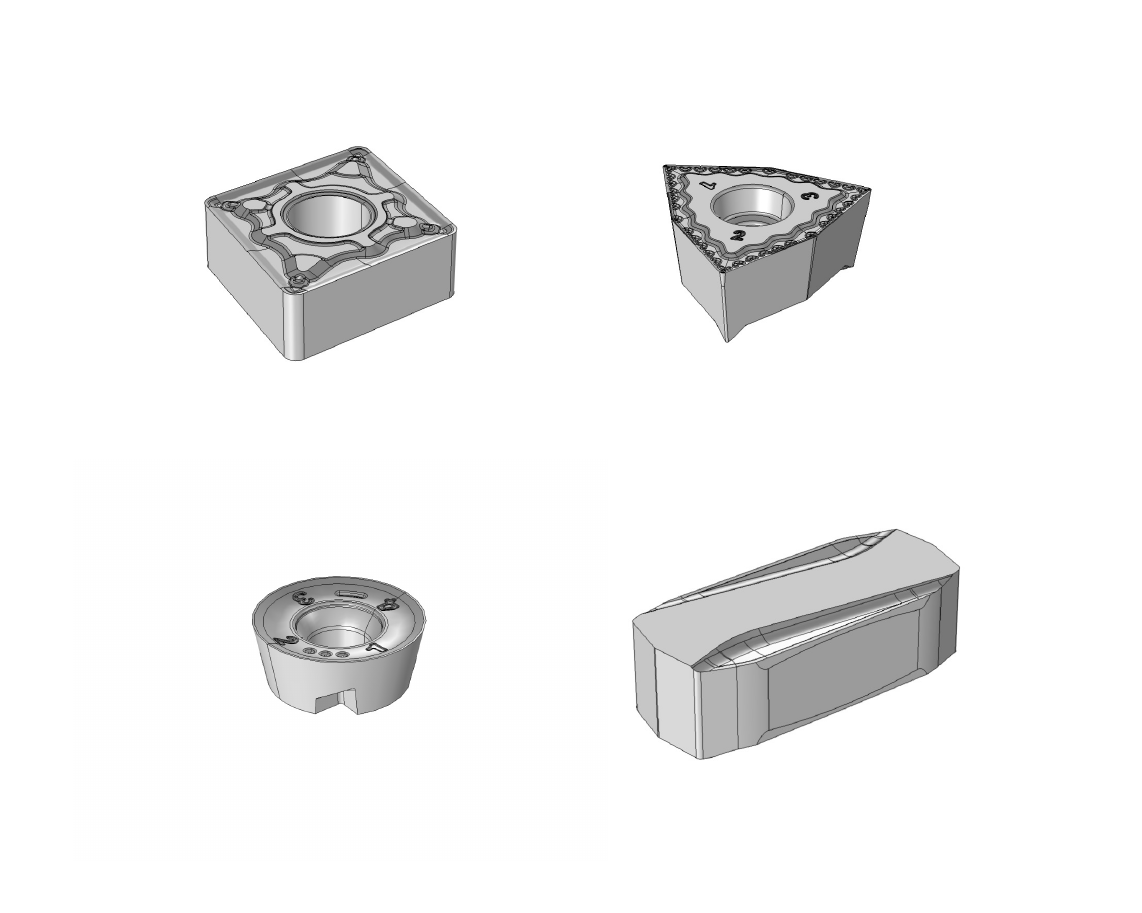}
   \caption{}
   \label{fig:inserts} 
\end{subfigure}
\hfill
\begin{subfigure}[b]{0.48\textwidth}
   \includegraphics[width=1\textwidth]{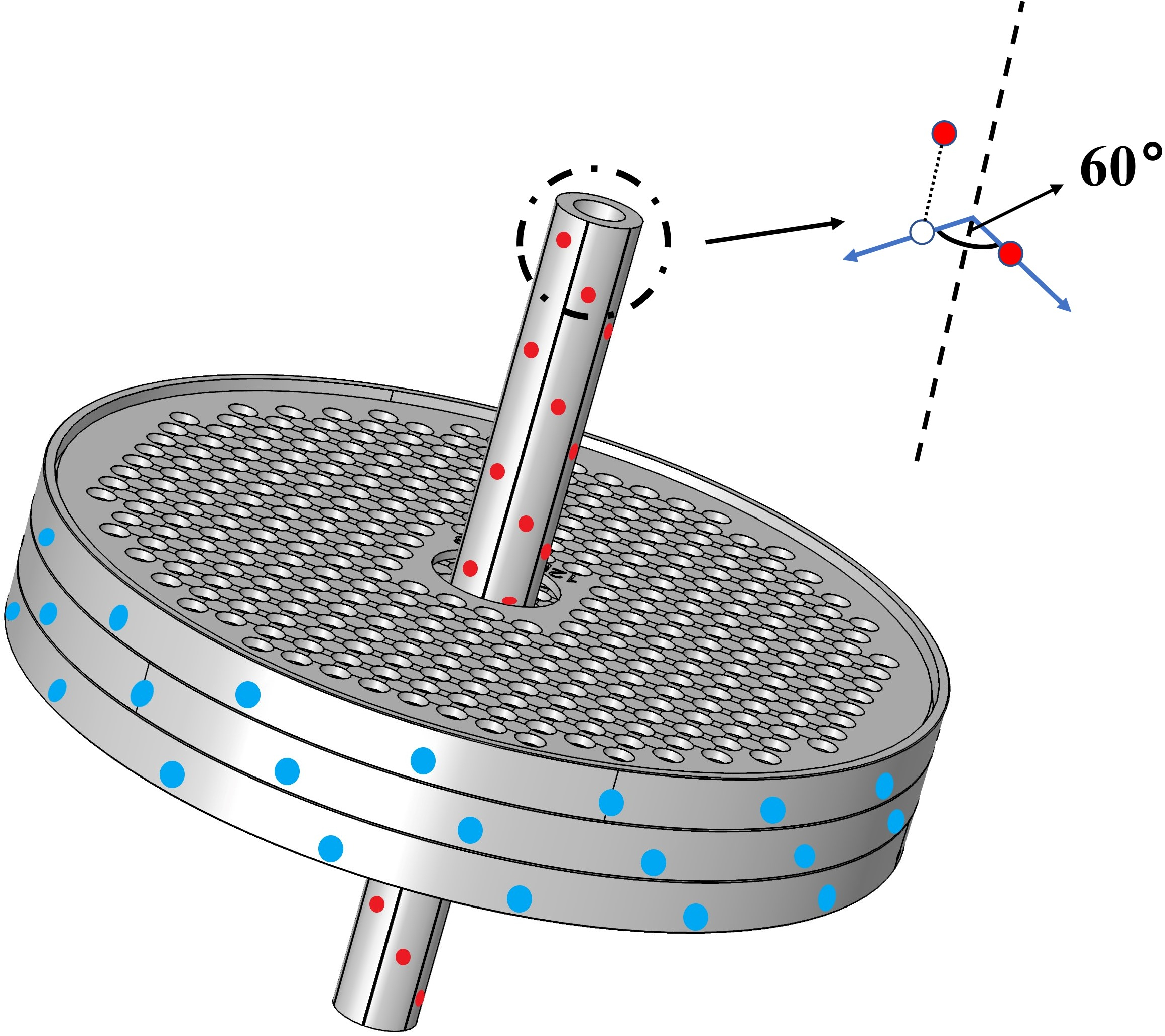}
   \caption{}
   \label{fig:3d-3disks}
\end{subfigure}
\caption{(a) Examples of the coated cutting tool inserts. (b) A 3D representation of a 3-disk part of the reactor. In red: inlet perforations on the rotating inlet tube. In blue: outlet perforations for each disk.}
\label{fig:geometry-explain}
\end{figure}

The studied process involves two coating steps carried out inside a commercial, industrial-scale Sucotec SCT600TH CVD reactor. To start with, a Ti(C,N) base layer of approximately 9 \textmu m is deposited on cemented carbide cutting tool inserts, shown in \cref{fig:inserts}. The second step involves the deposition of an alumina layer under specific conditions: $T$=1005\textdegree{}C and $p$=80 mbar, from a mixture of gas reactants that includes  AlCl\textsubscript{3}–CO\textsubscript{2}–HCl–H\textsubscript{2}–H\textsubscript{2}S. This step takes around 3 hours to complete \citep{hochauerCarbonDopedAAl2O32012}.

The CVD reactor consists of 40-50 perforated disks, stacked one on top of the other. The inserts to be coated are placed on each disk. For illustrative purposes, a schematic of three such disks is shown in \cref{fig:3d-3disks}. Specially designed perforations on a rotating cylindrical tube, which is placed in the center of the reactor, ensure the uniform distribution of the gas reactants over and around the inserts: the perforations are placed antipodally and there is a 60\textdegree{} angle difference between the axis connecting the inlets at each disk level. The feeding tube rotates at a fixed rotational speed of 2 RPM.

It is worth noting that each insert has a dedicated disk design which ultimately suggests that the interior geometry of the CVD reactor changes every time that it is set up.

The desired process outcome is uniform coating thickness distribution for the same insert and also uniform mean thickness across all production runs, all reactors and all production sites, as this ensures consistent product life (quality) \citep{bar-henExperimentalStudyEffect2017}. In practice, the desired uniformity is not always achieved, and therefore a systematic way of identifying the influential aspects of coating uniformity becomes necessary.

\subsection{Available data}
\label{sec:available-data}
\begin{figure}[p]
    \centering
    \includegraphics[width=\textwidth]{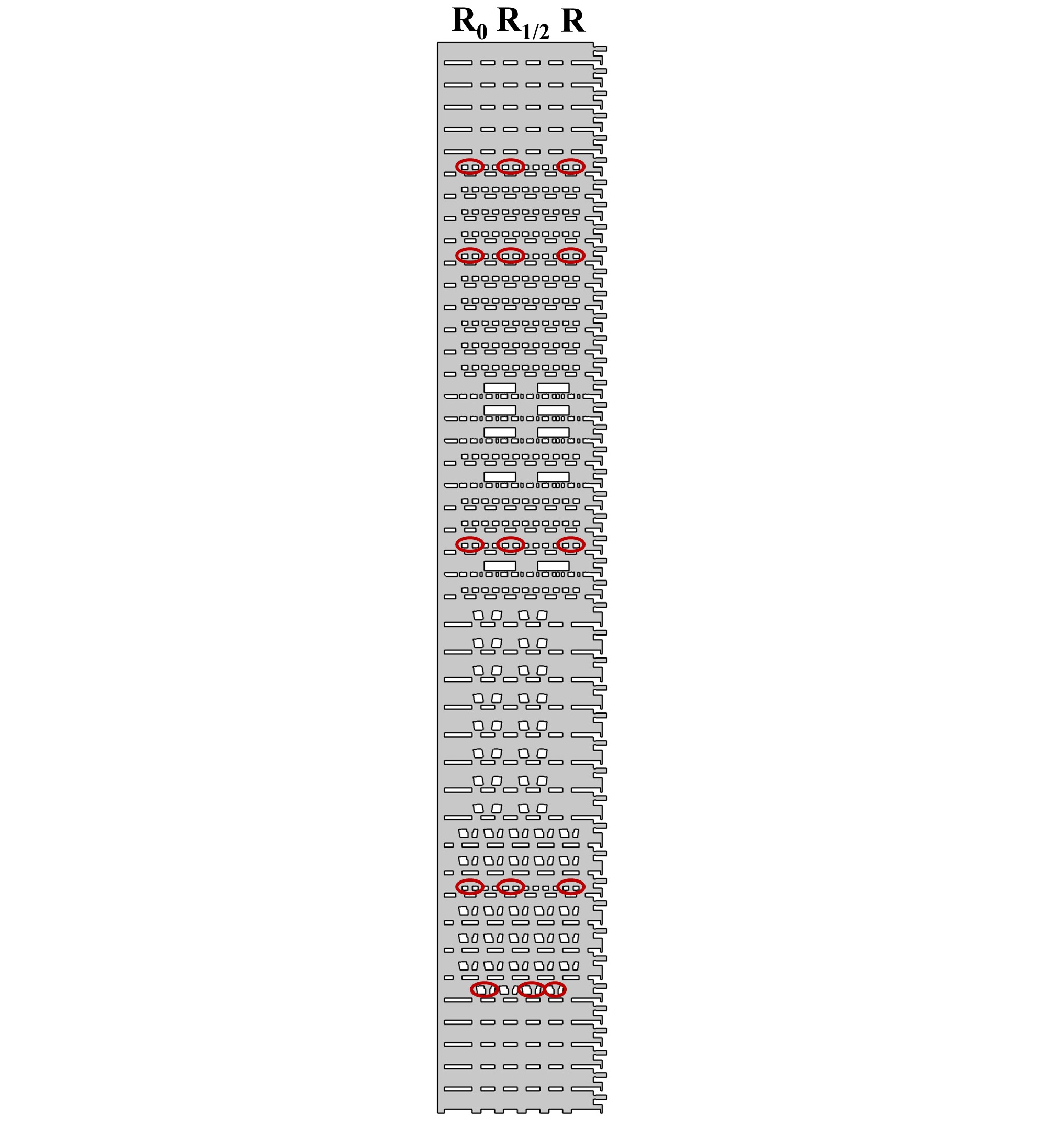}
    \caption{Most common measurement positions among the production data. These measurements can be used for several tasks, such as the development of CFD or ML approaches for the prediction of the process outcome.}
    \label{fig:measurement-positions}
\end{figure}

For each production run, thickness measurements are taken at three positions on five disks of interest, schematically shown in a representative geometry in \cref{fig:measurement-positions}. The thickness of the Ti(C,N) and \alumina{} coating layers is measured using the Calotest method \citep{lepickaInitialEvaluationPerformance2019}. These measurements have been utilized in previous work, both for the development of a CFD model of the process \citep{papavasileiouEfficientChemistryenhancedCFD2022}, and for the implementation of ML approaches for the prediction of coating thickness \citep{papavasileiouEquationbasedDatadrivenModeling2023}. 

Coating thickness is a vital measure of product quality for CVD applications. The long-term experience of the practitioners led to the selection of these 15 measurements for testing the quality of each production run. It should be noted that in case additional quality-related data (i.e. roughness of the coating) become available, they can be easily incorporated in the framework presented in this paper, in conjunction with thickness.


Additionally, the available dataset contains information about a) the process input parameters and b) the reactor geometry and setup. Some examples of these features include, but are not limited to:

\begin{itemize}
    \item The components of the reactor setup that determine the overall interior geometry, i.e. the sequence according to which the disk/inserts are stacked to form the overall reactor.
    \item The surface area of the inserts on each disk.
    \item The production \enquote{recipe}, a feature that encodes several process parameters and steps. We should note that there can be several versions of one recipe. There are a total of four base recipes present in the dataset with five versions for each (marked $V21$, $V20$, and older variants). This makes up a total of 20 recipes.
    \item The serial number of the reactor used for the production run.
\end{itemize}

An important contribution of this work emerged in the context of data exploration and pre-processing. It became necessary to engineer additional features, based on our intuition (subject matter expertise) regarding the existing inputs. These engineered features include the total surface area per reactor, the standard deviation of the surface area within the reactor, and the difference between the nominal and actual surface area within the reactor. The nominal surface area is the surface area considered by the production recipe and does not always coincide with the actual surface area. For more information on the available data, its type and its characteristics, the interested reader is referred to previous work by \citet{papavasileiouEquationbasedDatadrivenModeling2023}. A comparison of our approach with systematic methods for feature combinations, such as polynomial combination or even symbolic regression, is underway and out of the scope of this work.

\section{Machine learning methods}\label{sec:ml-tools}

\subsection{Unsupervised learning}
Unsupervised learning algorithms take unlabeled data as inputs to discover interesting patterns in the data (e.g., association rule analysis) or try to create subgroups - or clusters - of similar observations within the dataset \citep{hastieUnsupervisedLearning2009}. Dimensionality reduction techniques such as the widely used Principal Component Analysis (PCA), autoencoders \citep{wangAutoencoderBasedDimensionality2016}, and diffusion maps \citep{koronakiDatadrivenReducedorderModel2020,koronakiPartialDataOutofsample2023a} also fall under unsupervised learning as they provide a reduced data representation without requiring the corresponding response. The clustering and dimensionality reduction techniques implemented are briefly discussed in the following sections.

\subsubsection{Clustering}\label{sec:clustering}
Clustering algorithms are based on the concept of dissimilarity (or similarity) between observations, which determines their grouping. Typically, these algorithms utilize a similarity matrix, where pairwise similarities between observations are represented. For quantitative variables, the commonly employed metric is the Euclidean distance,  while alternative distance metrics can also be used \citep{hastieUnsupervisedLearning2009,jamesUnsupervisedLearning2021}.

Clustering algorithms are categorized into various categories. Partitional approaches, such as the k-means algorithm, involve assigning observations to clusters based on distances to centroids iteratively, requiring an \textit{a priori} choice of the number of clusters and sensitive to initial centroid positions \citep{macqueenMethodsClassificationAnalysis1967}. Density-based algorithms, such as OPTICS and DBSCAN, identify clusters by considering areas of high density separated by low-density regions. Certain algorithm parameters, such as the minimum points that form a cluster and the minimum distance between the core points require specification \citep{ankerstOPTICSOrderingPoints1999,esterDensitybasedAlgorithmDiscovering1996,schubertDBSCANRevisitedRevisited2017}. Hierarchical clustering methods link data points according to criteria, progressively creating clusters until a single cluster is achieved in the case of agglomerative clustering, or progressively splitting clusters starting until each observation is its own cluster in the case of divisive clustering. The results depend on the distance metric and the linkage criteria selected \citep{murtaghAlgorithmsHierarchicalClustering2012,vijayaComparativeStudySingle2019}. Additional methods include model-based and spectral methods \citep{fraleyModelBasedClusteringDiscriminant2002,jiaLatestResearchProgress2014}.

Here, we focus on \textit{agglomerative} hierarchical clustering, implementing a Ward linkage criterion for merging the clusters. This is an established variance minimization approach \citep{wardHierarchicalGroupingOptimize1963} that works by minimizing the sum of squared differences within all clusters. Agglomerative hierarchical clustering is selected because it provides insight on how the data merges depending on the number of clusters chosen. This information is readily available in the form of a dendrogram, such as the one presented in \cref{fig:dendro}.

For this specific problem, the 15 available thickness measurements of 603 production runs are used as inputs (cf. \cref{sec:available-data}). The clustering results are then interpreted based on the characteristics of the resulting clusters. Our goal is to identify production runs that are similar to each other and to try to uncover the discerning features of these clusters.

\subsection{Supervised learning}\label{sec:supervised-learning}

Supervised learning algorithms, unlike unsupervised ones, require labeled data, associating features ($x_i$) with responses ($y_i$). Supervised learning tasks include regression for continuous variables and classification for binary or ordinal responses \citep{jamesStatisticalLearning2021}.

The methods evaluated for this work include: (a) linear methods: for regression, lasso \citep{tibshiraniRegressionShrinkageSelection1996}, and ridge \citep{hoerlRidgeRegressionBiased1970} regression and logistic regression for classification tasks. (b) Support vector machines (SVMs) \citep{cortesSupportvectorNetworks1995} that can be categorized as linear or nonlinear methods based on the kernel used for classification tasks. (c) Tree-based methods: involving classification and regression trees \citep{breimanClassificationRegressionTrees1984} and their ensemble counterparts such as random forests \citep{breimanRandomForests2001}, gradient-boosted trees \citep{friedmanGreedyFunctionApproximation2001},  extra trees \citep{geurtsExtremelyRandomizedTrees2006},  and XGBoost \citep{chenXGBoostScalableTree2016}, which combine numerous trees to enhance performance \citep{hastieEnsembleLearning2009}. (d) Artificial neural networks (ANN), whose diverse architectures \citep{aggarwalNeuralNetworksDeep2018} can provide valuable options for both classification and regression tasks.

In this work, logistic regression, random forests, SVM, extra trees, gradient-boosted trees, XGBoost, and ANNs are implemented for supervised learning tasks. However, only the methods that demonstrate the best performance for our dataset are presented in \cref{sec:results}.

\subsection{Shapley values}
Shapley values, originally introduced by \citet{shapleyValueNPersonGames1952} and proposed as a tool to analyze machine learning models in \citep{lundbergUnifiedApproachInterpreting2017, lundbergLocalExplanationsGlobal2020} intricately assess the average contribution of each feature's value to predictions, providing an understanding of how alterations to a variable might influence the ultimate model output.

In the context of this work, a SHAP (Shapley value based) analysis is conducted on the proposed regression models (cf. \cref{sec:supervised-learning,sec:shapley-analysis}) and the resulting Shapley values will shed light on the importance of each feature to the model output.

\section{Results}
\label{sec:results}
\subsection{Clustering}
As mentioned in \cref{sec:clustering}, the agglomerative hierarchical clustering algorithm with a Ward linkage criterion is implemented for clustering the 603 production runs.


The clustering algorithm utilizes the 15 thickness measurements for each of the 603 production runs, forming a $603\times15$ matrix. Clusters are then created solely based on the process outputs. Subsequently, the distinctive features are identified by analyzing the process inputs for each production run.

\begin{figure}[h!]
\captionsetup[subfigure]{justification=centering}
\centering
\begin{subfigure}[b]{0.49\textwidth}
   \includegraphics[width=1\textwidth]{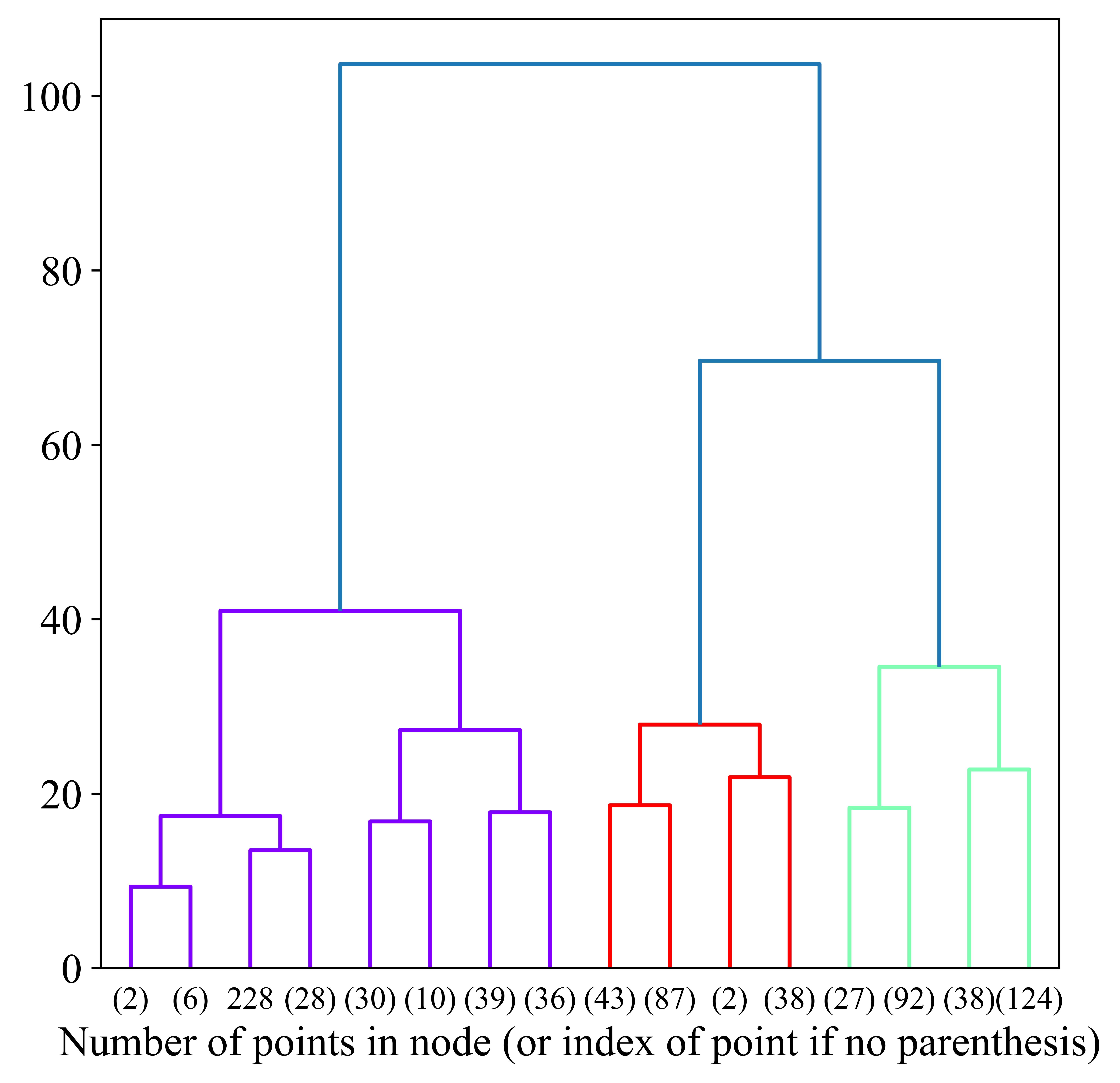}
   \caption{}
   \label{fig:dendro}
\end{subfigure}
\begin{subfigure}[b]{0.49\textwidth}
   \includegraphics[width=1\textwidth]{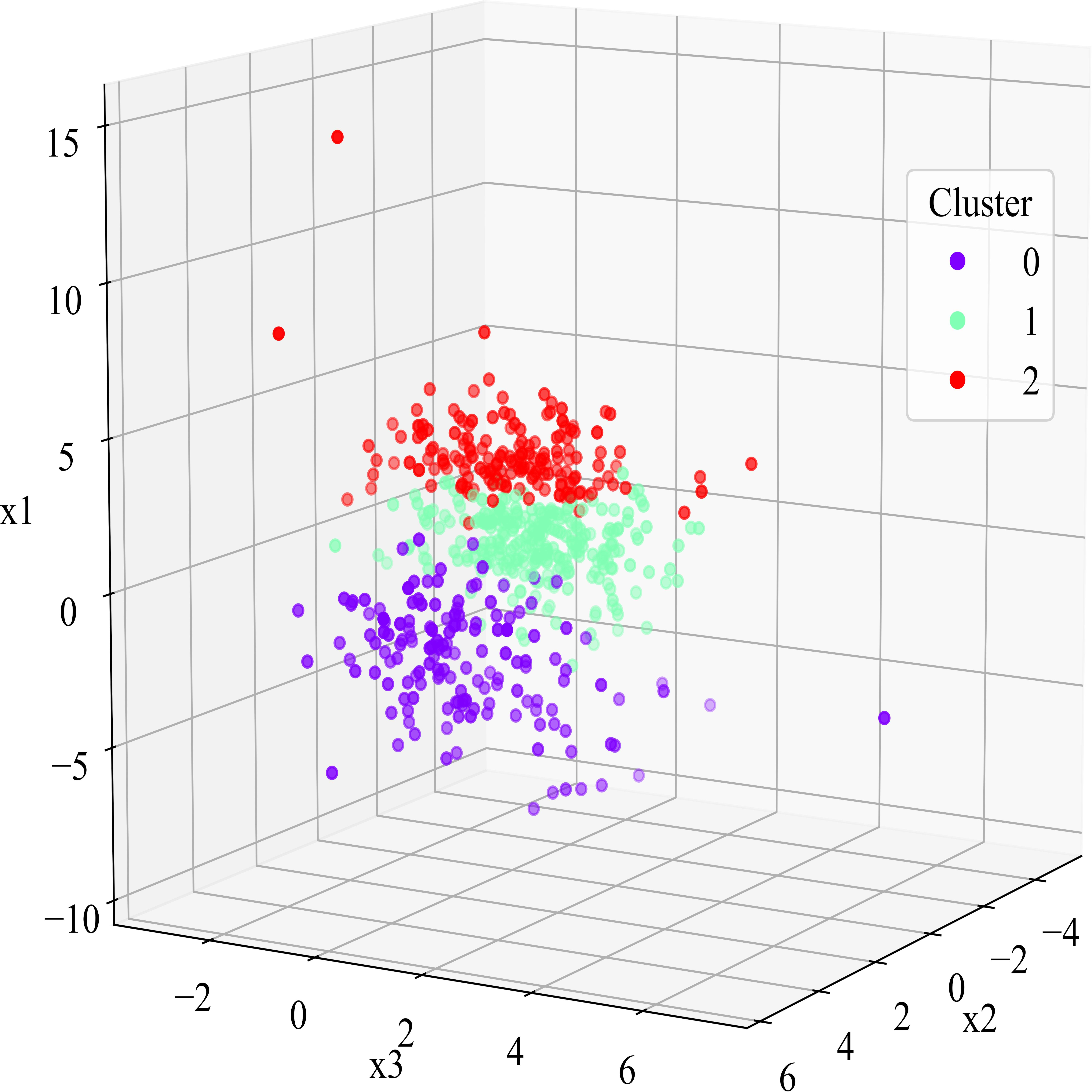}
   \caption{}
   \label{fig:clusters-visual} 
\end{subfigure}
\hfill
\caption{(a) Resulting dendrogram of the clusters output by the implemented agglomerative hierarchical clustering algorithm using a Ward linkage criterion. The three main clusters of interest are colored purple, red, and green. We note that by selecting a slightly higher dissimilarity threshold, the red and green clusters can be merged and viewed as a larger cluster (shown in blue). (b) The three resulting clusters, visualized in a reduced 3D space. The three clusters appear to be well-formed. PCA was used for finding the 3D reduced space.}
\label{fig:pca-dendro}
\end{figure}

The hierarchical clustering algorithm generates a dendrogram that illustrates cluster levels, member counts, and dissimilarities. The clusters are depicted as branches of a tree, culminating in the "trunk," representing the final cluster (by agglomerating smaller ones). In our case, the resulting dendrogram is shown in \cref{fig:dendro}. By selecting a dissimilarity threshold, we can discern one, two, three, or more clusters. In \cref{fig:dendro}, the three clusters are colored purple, red, and green. A higher dissimilarity threshold merges the red and green clusters into a single blue cluster (as shown in \cref{fig:dendro}). The resulting clusters are visualized in a reduced three-dimensional space (through projection on three principal components) in \cref{fig:clusters-visual}.

\begin{figure}[h!]
\captionsetup[subfigure]{justification=centering}
\centering
\begin{subfigure}[b]{0.49\textwidth}
       \includegraphics[width=1\textwidth]{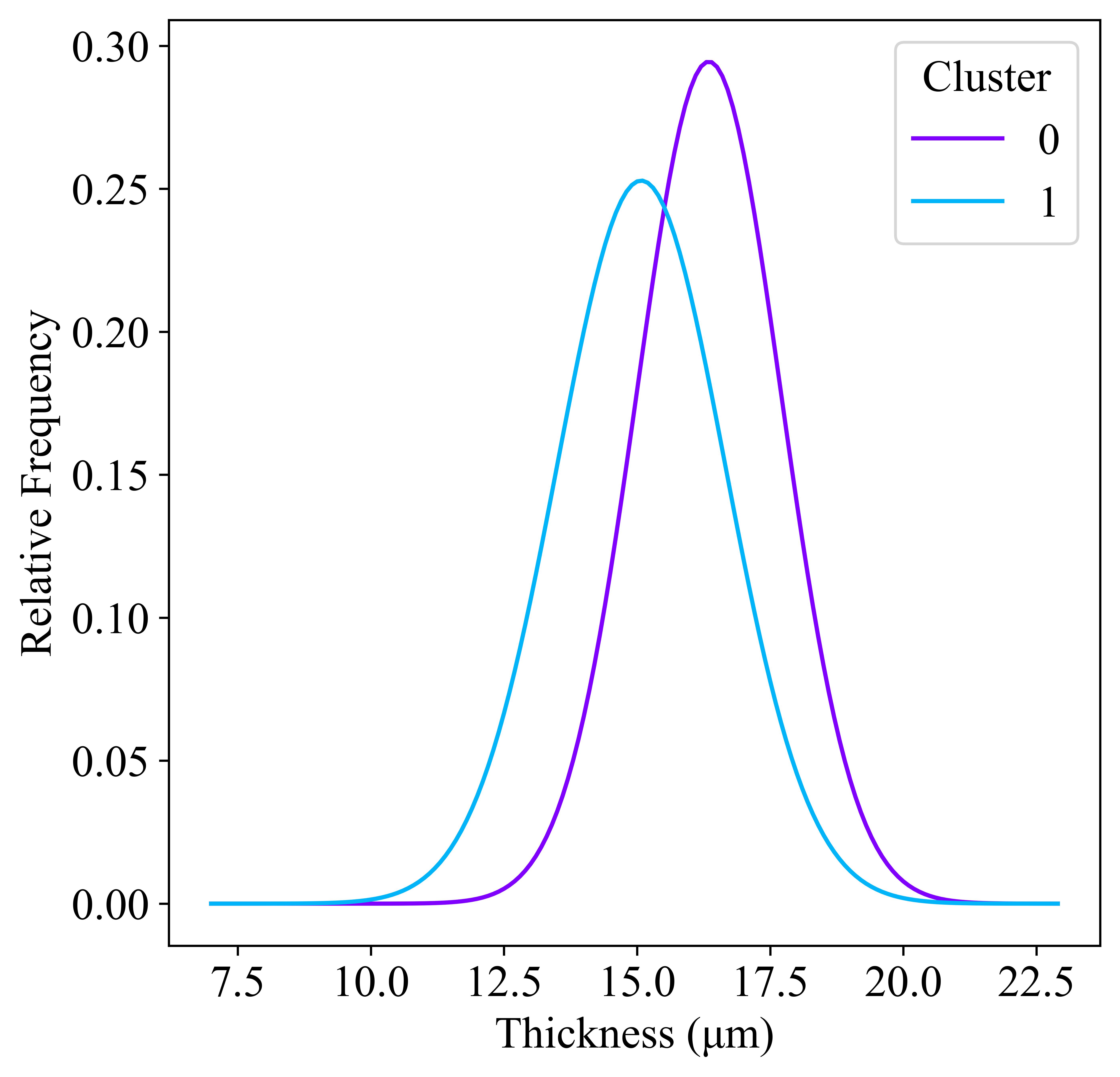}
   \caption{}
   \label{fig:dist-2-clust} 
\end{subfigure}
\hfill
\begin{subfigure}[b]{0.49\textwidth}
   \includegraphics[width=1\textwidth]{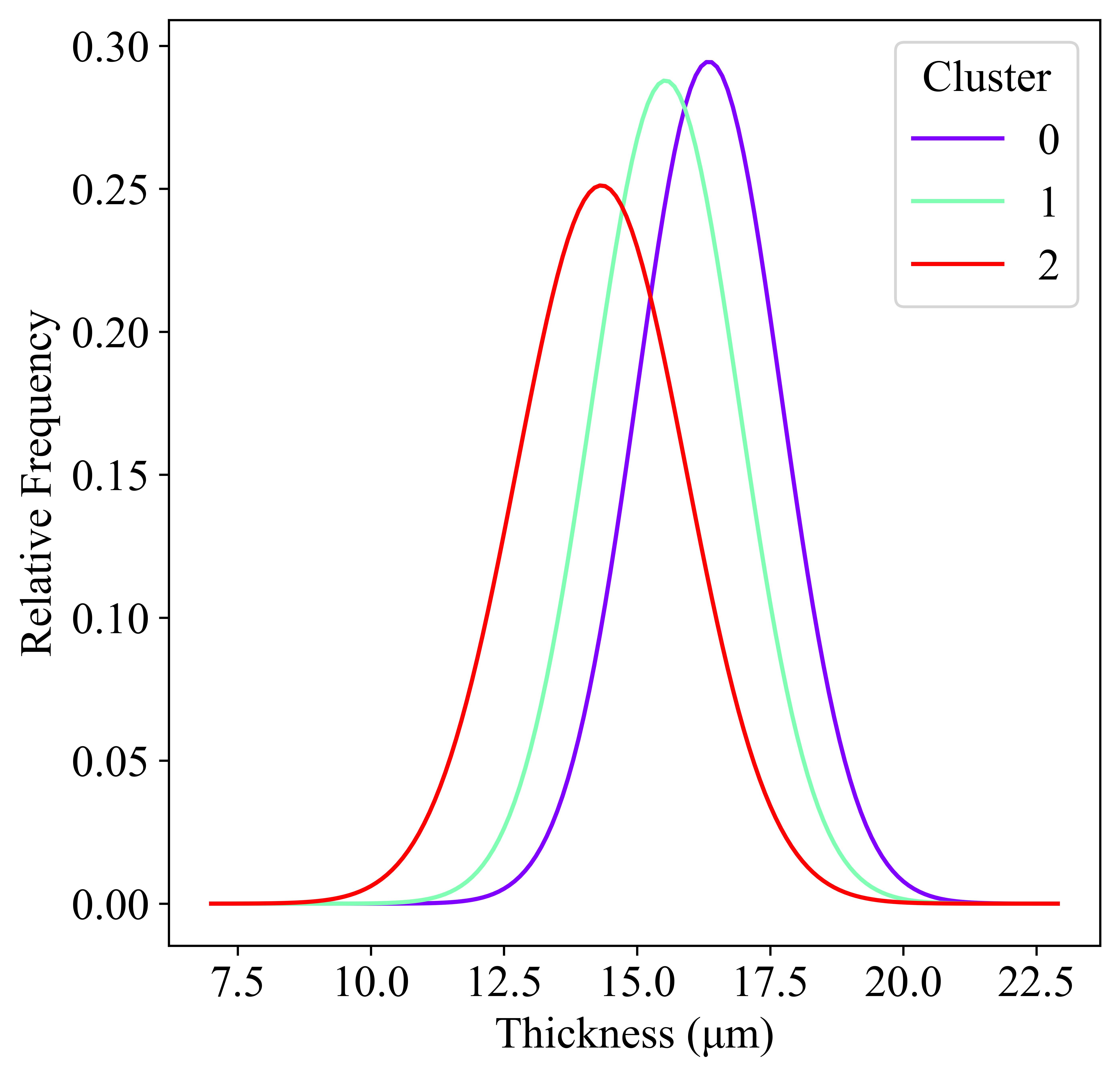}
   \caption{}
   \label{fig:dist-3-clust}
\end{subfigure}
\caption{Thickness distribution in the case of: a) 2 clusters and b) 3 clusters. High average thickness and low standard deviation is a measure of process efficiency and product quality. The production runs in the \enquote{purple} cluster demonstrate superior quality characteristics.}
\label{fig:dist-clust}
\end{figure}

As mentioned above, the thickness and its uniformity throughout production runs is a very effective process performance indicator and product quality metric. Thus, production runs with a higher average thickness and a lower standard deviation can be considered superior to those with a lower average thickness and higher standard deviation.  We observe that the thickness within the clusters follows a normal distribution, and therefore we can calculate the first and second statistical moments (that is, the mean ($\mu_\mathrm{thick}$), and standard deviation ($\sigma_\mathrm{thick}$) and visualize the thickness distributions as shown in \cref{fig:dist-clust}. 

\subsection{Critical input identification}
\label{sec:crit-inputs}

In this section, the focus shifts to the process inputs whose variation is critical for each cluster. We propose three different ways for assessing the relative importance of process inputs.
\begin{enumerate}
    \item Intuition-based approach: By finding characteristics that are predominantly different in each cluster, we can assess their importance on the process outcome (cf. \cref{sec:crit-inputs-one}).    
    \item Supervised learning approach: Classification algorithms are trained using the cluster labels of the clustering step as outputs and various inputs: some process inputs lead to higher accuracy, which is an indication of their importance. Conversely, less important inputs have an adverse effect on the accuracy of the classifier (cf. \cref{sec:classification}).
    \item Shapley value approach: The importance of input features for classification or regression can be assessed using Shapley values (cf. \cref{sec:shapley-analysis}).
\end{enumerate}

\subsubsection{Combining clustering and subject matter expertise}
\label{sec:crit-inputs-one}

When two clusters are considered (\cref{fig:dist-2-clust}), cluster 0 demonstrates superior characteristics, with the highest average thickness and the lowest standard deviation (\cref{tab:2clust}). Further examination reveals that cluster 0 is characterized by production runs predominantly using recipe version $V21$, while cluster 1 comprises runs using version $V20$ and older versions, indicating recipe version as the main distinguishing feature.

\begin{table}[h!]
\centering
\caption{Characteristics of each cluster in the case of two clusters. The recipe version used for production is the discerning feature of the two clusters.}
\begin{adjustbox}{width=1\textwidth}
\begin{tabular}{c c c c}
\hline
\textbf{Cluster} & \textbf{ $\boldsymbol{\mu_\mathrm{thick}}$ (\textmu m)} & \textbf{$\boldsymbol{\sigma_\mathrm{thick}}$ (\textmu m)}  & \textbf{Predominant recipe versions} \\ 
\hline
0 & 16.35  & 1.355  & $V21$          \\
1  & 15.08  & 1.578  & $V20$ \& older \\
\hline
\end{tabular}
\label{tab:2clust}
\end{adjustbox}
\end{table}

When three clusters (\cref{fig:dist-3-clust}), are identified by the clustering algorithm, cluster 0 exhibits superior characteristics, with the highest average thickness and the lowest standard deviation (\cref{tab:3clust}). In particular, cluster 0 comprises production runs using recipe version $V21$, and it is practically the same as cluster 0 in the two-cluster case mentioned in the previous paragraph. Clusters 1 and 2 predominantly use $V20$ and older versions and are the result of the splitting of cluster 1 identified in the two-cluster case. This cluster splitting, in essence, means that even among production runs using recipe version $V20$ and older, there are certain cases where favorable quality characteristics are achieved. This raises the question: which is the critical input that led to this difference in quality?

\begin{figure}[h!]
\captionsetup[subfigure]{justification=centering}
\centering
\begin{subfigure}[b]{0.4\textwidth}
   \includegraphics[width=1\textwidth]{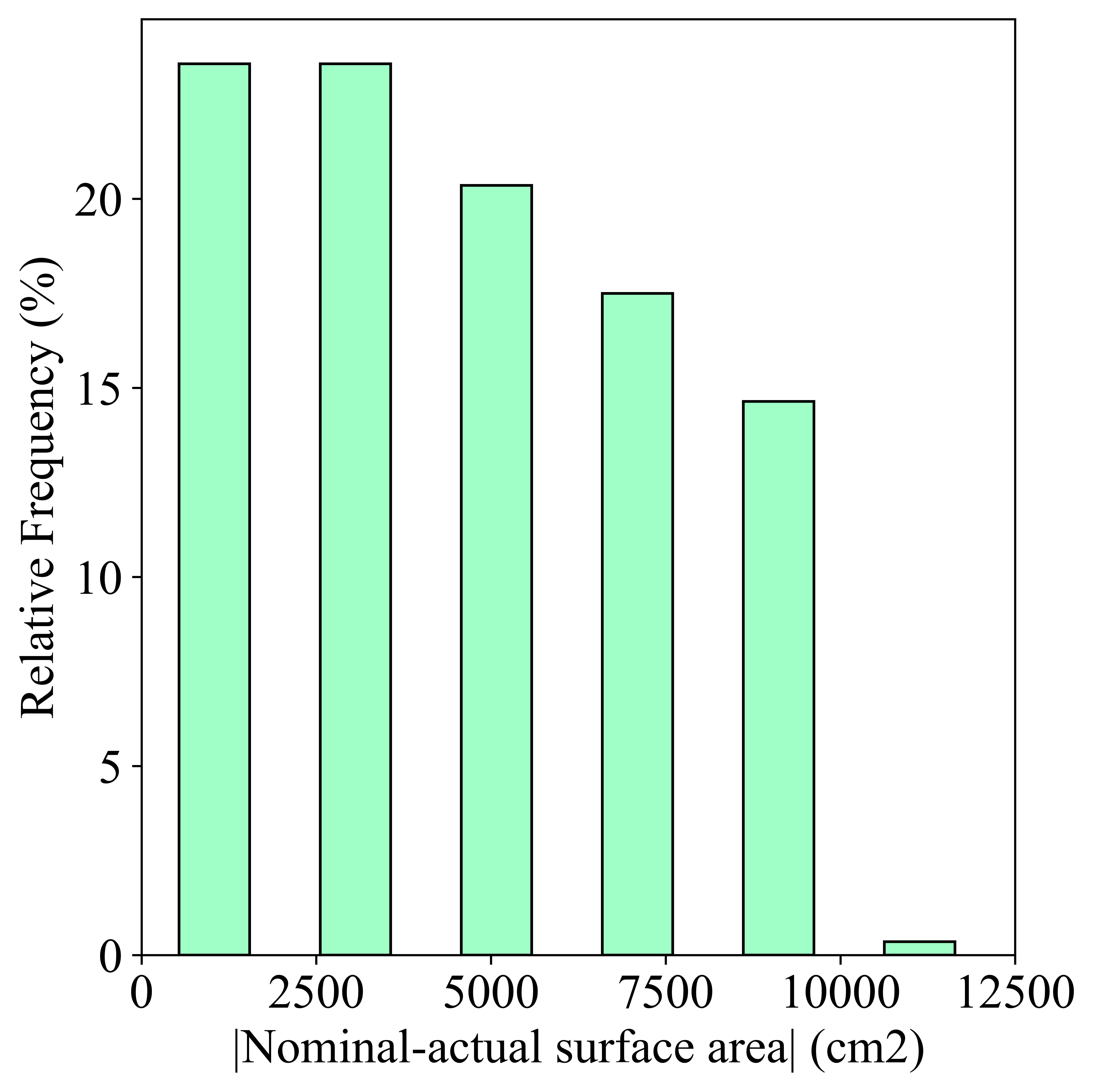}
   \caption{}
   \label{fig:metric-clust-1} 
\end{subfigure}
\begin{subfigure}[b]{0.4\textwidth}
   \includegraphics[width=1\textwidth]{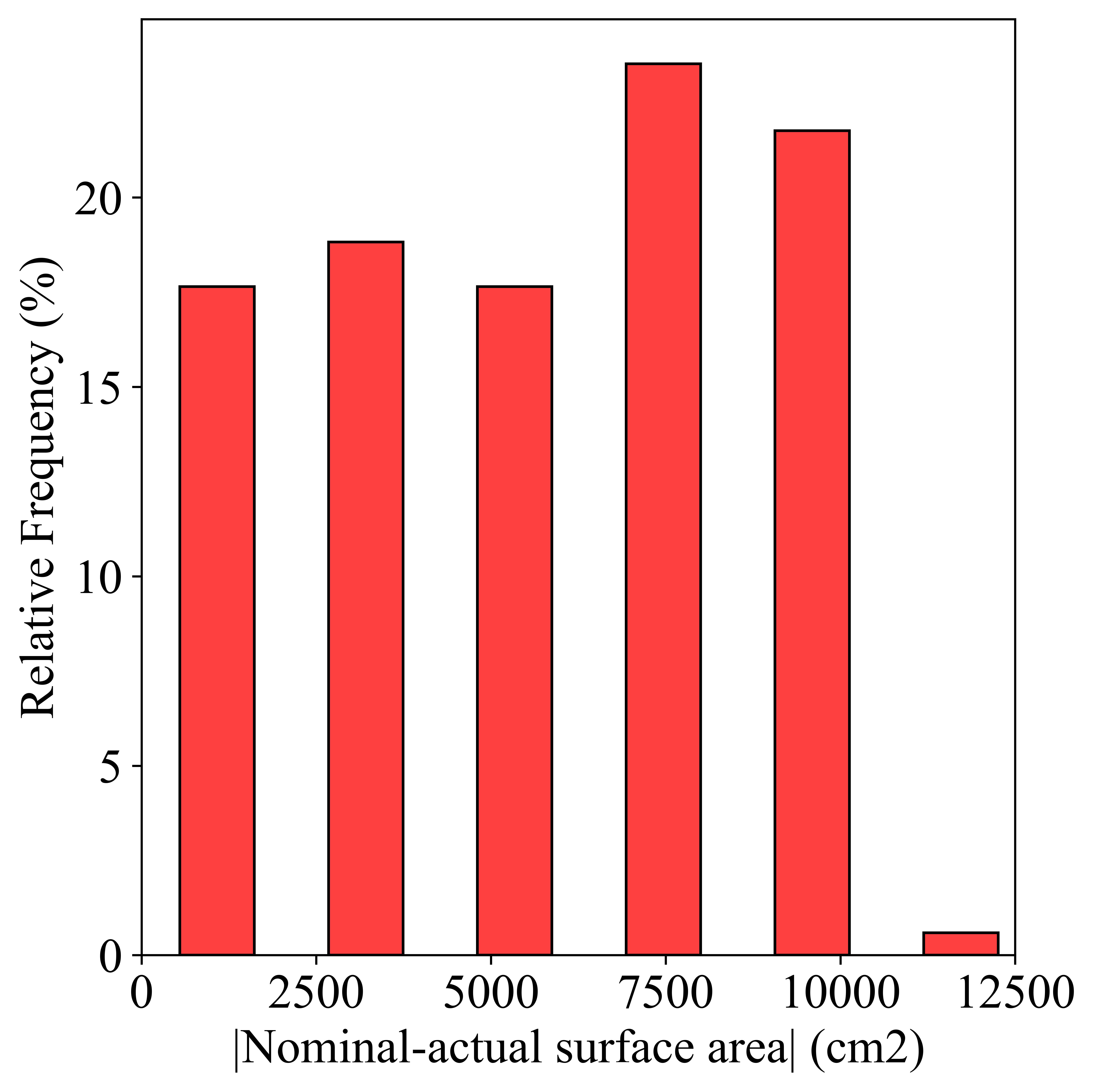}
   \caption{}
   \label{fig:metric-clust-2}
\end{subfigure}
\caption{Distributions of |Nominal recipe surface area - actual surface area| for clusters 1 (in green) and 2 (in red). Cluster 2 includes relatively more observations with values larger than 5000 cm\textsuperscript{2} when compared with cluster 1.}
\label{fig:dist-metric}
\end{figure}

Further assessment drew our attention to an engineered feature, the absolute value of the difference between the nominal and actual total surface area to be coated. The nominal surface area is the predetermined production setting, specified for increments of $1 m^2$. In practice, this rarely matches the actual total surface area value and this discrepancy is evident when comparing the distributions between clusters 1 and 2, as shown in \cref{fig:dist-metric}; On average, for the members of cluster 2, the difference between the nominal and actual total surface area is greater than $0.5 m^2$, while in cluster 1 it is less than $0.5 m^2$. This analysis suggests that when the value of this difference is less than $0.5 m^2$, the qualitative characteristics of the products are superior, thus leading to a clear and cost-free suggestion for improvement: define preset production parameters for increments of $0.5 m^2$ (instead of $1 m^2$) of the total surface area. 

\begin{table}[h!]
\centering
\caption{Characteristics of three clusters: Discerning features include the recipe version used for production and the absolute difference between nominal and actual surface area.}
\begin{adjustbox}{width=1\textwidth}
\begin{tabular}{c c c c c}
\hline
\textbf{Cluster} & \textbf{ $\boldsymbol{\mu_\mathrm{thick}}$ (\textmu m)} & \textbf{$\boldsymbol{\sigma_\mathrm{thick}}$ (\textmu m)} & \textbf{Predominant recipe versions} & \begin{tabular}[c]{@{}c@{}} \textbf{|Nominal recipe surface area} \\ \textbf{- actual surface area| (cm\textsuperscript{2})}\end{tabular}  \\
\hline
0  & 16.35  & 1.354  & $V21$          &  4892  \\
1  & 15.53  & 1.386  & $V20$ \& older &  4628  \\
2  & 14.32  & 1.588  & $V20$ \& older &  5526  \\
\hline
\end{tabular}
\label{tab:3clust}
\end{adjustbox}
\end{table}

\subsection{Classification}
\label{sec:classification}

We train a classifier to predict cluster labels that resulted from the clustering analysis, using as inputs the dominant features identified in the previous section (clustering). This is useful in practice to predict the overall quality characteristics of the production run, as these cluster labels correspond to distinct thickness distributions.

The results for a binary (two-cluster case) and a multi-label classification (three-cluster case) task are presented. For these tasks, we divide the 603 observations into a training set and a test set using an 80/20 ratio.

Initially, classification models take as input the two important features identified through clustering. However, these are not the only discernible differences between clusters; other features, such as the year of production, the reactor used, and the standard deviation of the surface area within the reactor, also have marked differences among clusters. Therefore, these inputs are also considered when training the classifier.

\begin{figure}[!ht]
\captionsetup[subfigure]{justification=centering}
\centering
\begin{subfigure}[b]{0.33\textwidth}
   \centering
    \includegraphics[width=\textwidth]{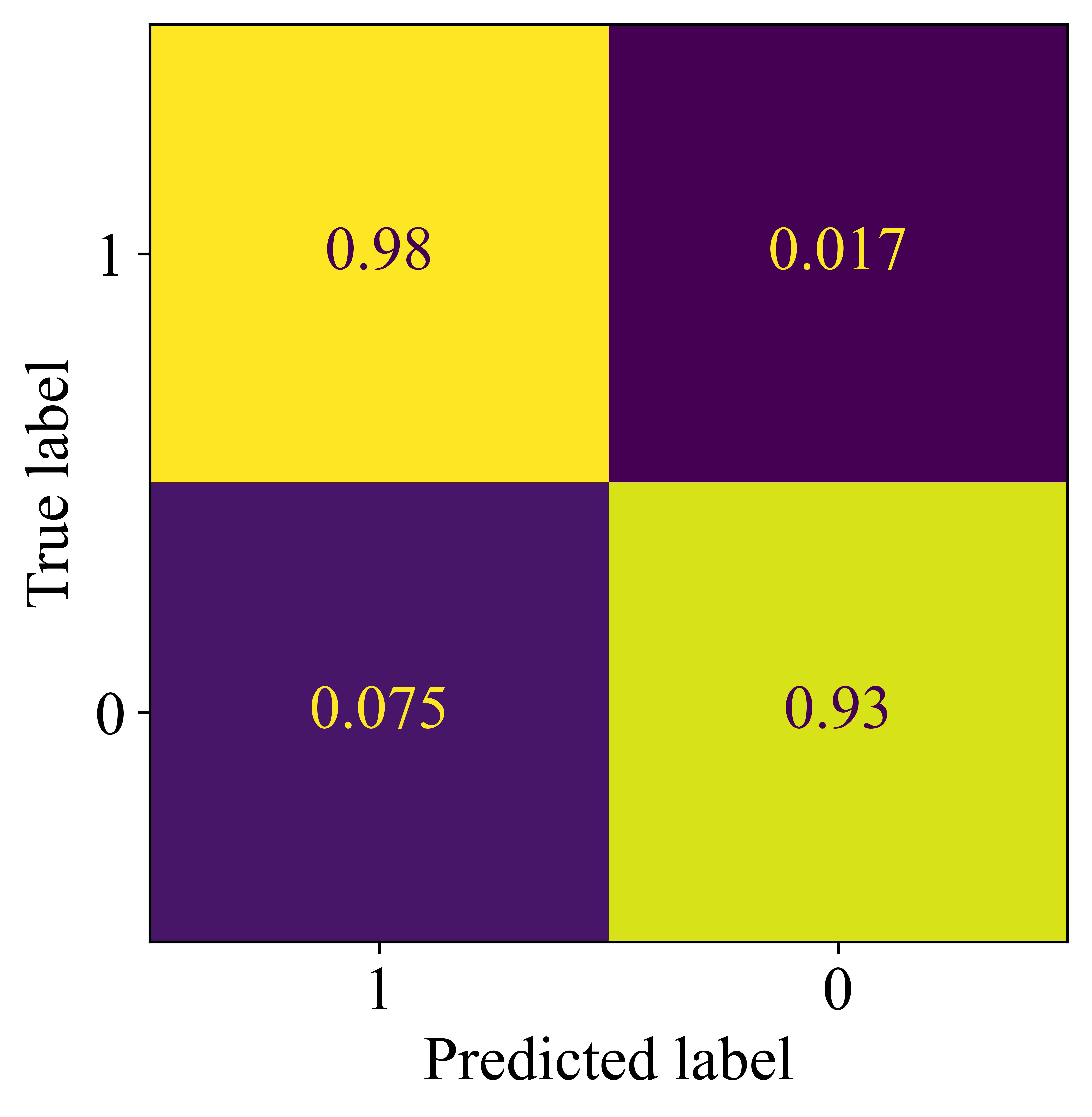}
    \caption{}
    \label{fig:cm-global-2-train}
\end{subfigure}
\begin{subfigure}[b]{0.33\textwidth}
   \centering
    \includegraphics[width=\textwidth]{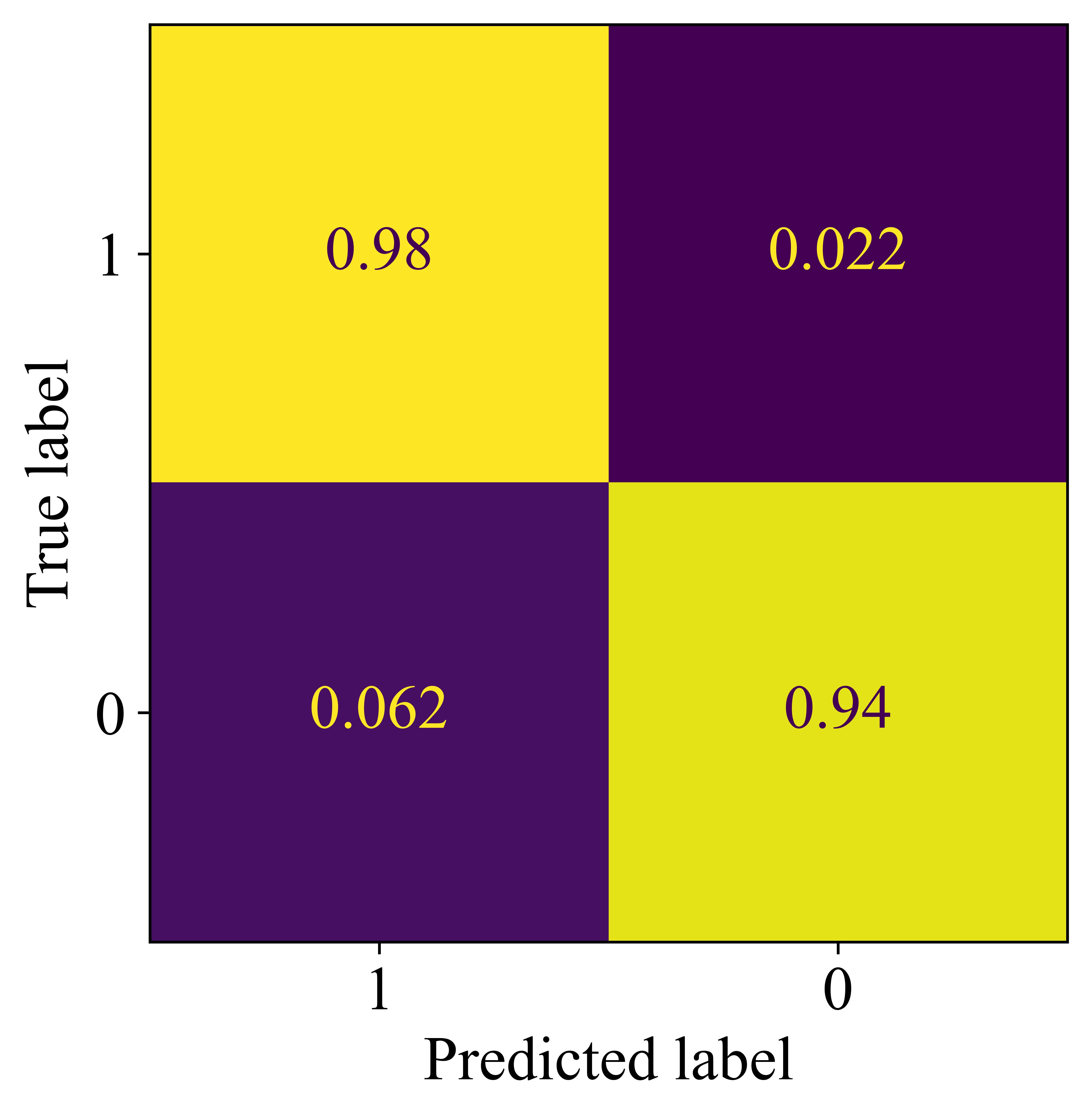}
    \caption{}
    \label{fig:cm-global-2-test}
\end{subfigure}
\begin{subfigure}[b]{0.33\textwidth}
\centering
\includegraphics[width=1\textwidth]{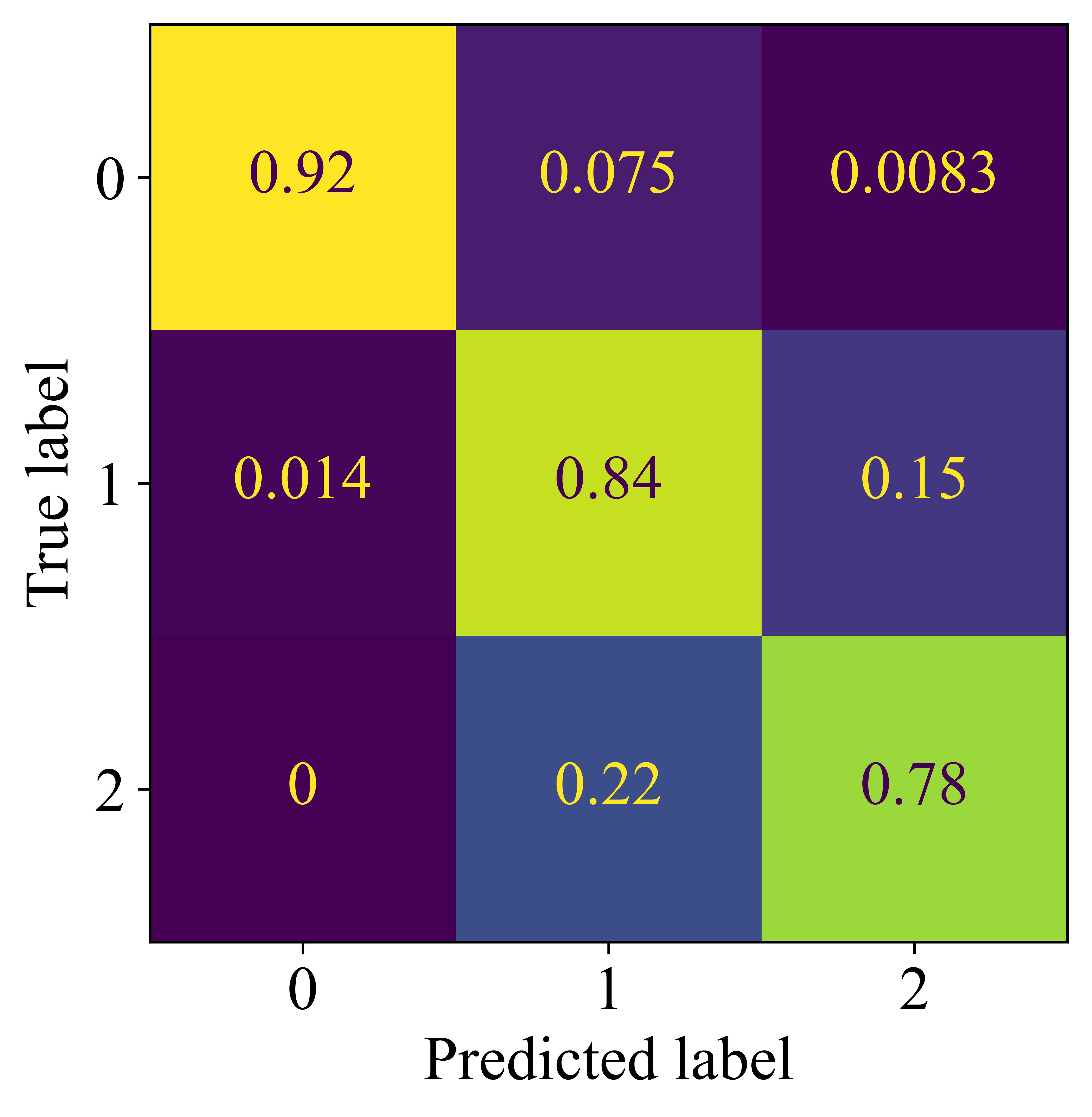}
   \caption{}
   \label{fig:cm-global-3-train}
\end{subfigure}
\begin{subfigure}[b]{0.33\textwidth}
\centering
\includegraphics[width=1\textwidth]{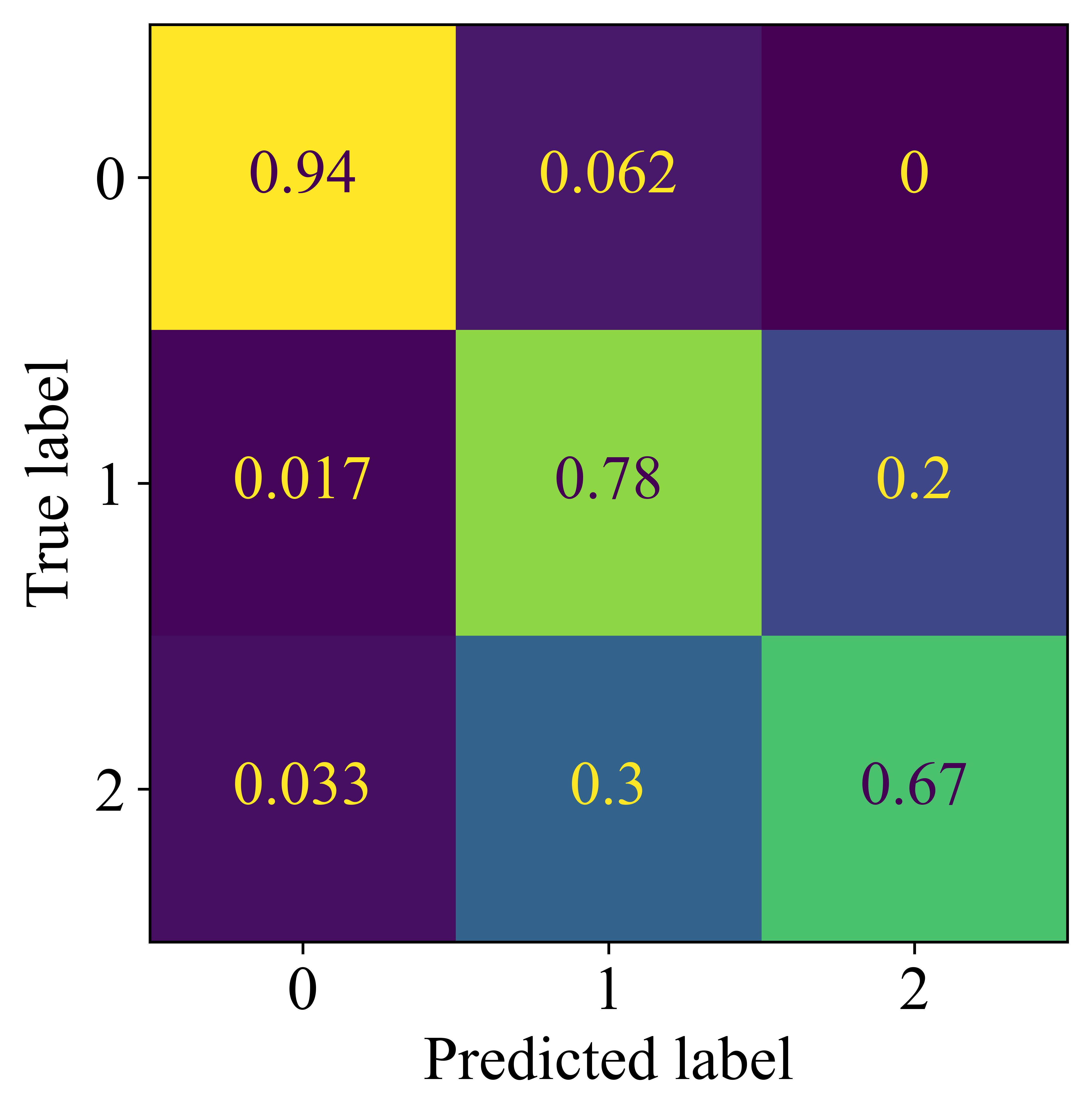}
    \caption{}
    \label{fig:cm-global-3-test}
\end{subfigure}
\caption{Confusion matrices for (a),(c) the training set and (b),(d) the test set of the two-cluster and three-cluster classification cases, respectively.}
\label{fig:global-class}
\end{figure}

The initial step involves training a random forest classification model (\textit{n\_estimators=1000, max\_depth=6}) to predict whether a production run belongs to cluster 0 or 1 in \cref{fig:dist-2-clust}, treating it as a binary classification problem. The classifier, as shown in the confusion matrices in \cref{fig:cm-global-2-train,fig:cm-global-2-test}, accurately distinguishes between clusters 0 and 1 production runs both for the training ($\mathrm{accuracy}=0.954$) and test set ($\mathrm{accuracy}=0.958$). The calculated accuracy, $f1$, precision, and recall metrics are presented in detail in \cref{tab:class-metrics}.

Subsequently, a random forest classification model (\textit{n\_estimators=1000, max\_depth=6}) is developed to determine if a production run belongs to cluster 0, 1, or 2 in \cref{fig:dist-3-clust}, making it a multi-label classification problem. As demonstrated in the confusion matrices in \cref{fig:cm-global-3-train,fig:cm-global-3-test}, the classifier identifies cluster 0 members very accurately, for both training and test datasets. However, it sometimes struggles to distinguish between members of cluster 1 and cluster 2, often misclassifying them as members of the other cluster. The accuracy of the classifier on the test set is 0.793. As in the two-cluster case, all metrics are presented in \cref{tab:class-metrics}. Since this is not a binary classification problem, the $f1$, precision and recall metrics are macro-averaged \citep{zhangReviewMultiLabelLearning2014}.

\begin{table}[h!]
\centering
\caption{Classification metrics for the two-cluster and three-cluster cases. The metrics for the three-cluster case have been macro-averaged.}
\begin{tabular}{l c c c c}
                 & Accuracy    & $f1$       & Precision    & Recall    \\
\hline
\multicolumn{5}{c}{\textbf{2-cluster case}}                          \\
\hline
Training Set     & 0.968       & 0.958    & 0.954        & 0.962     \\
Test Set         & 0.967       & 0.958    & 0.958        & 0.958     \\
\hline
\multicolumn{5}{c}{\textbf{3-cluster case (macro-averaged metrics)}} \\
\hline
Training Set     & 0.840       & 0.848    & 0.844        & 0.852     \\
Test Set         & 0.793       & 0.792    & 0.795       & 0.790    \\
\hline
\end{tabular}
\label{tab:class-metrics}
\end{table}

\subsection{Regression}
\label{sec:reg-results}
\begin{figure}[h!]
\centering
\includegraphics[width=0.50\textwidth]{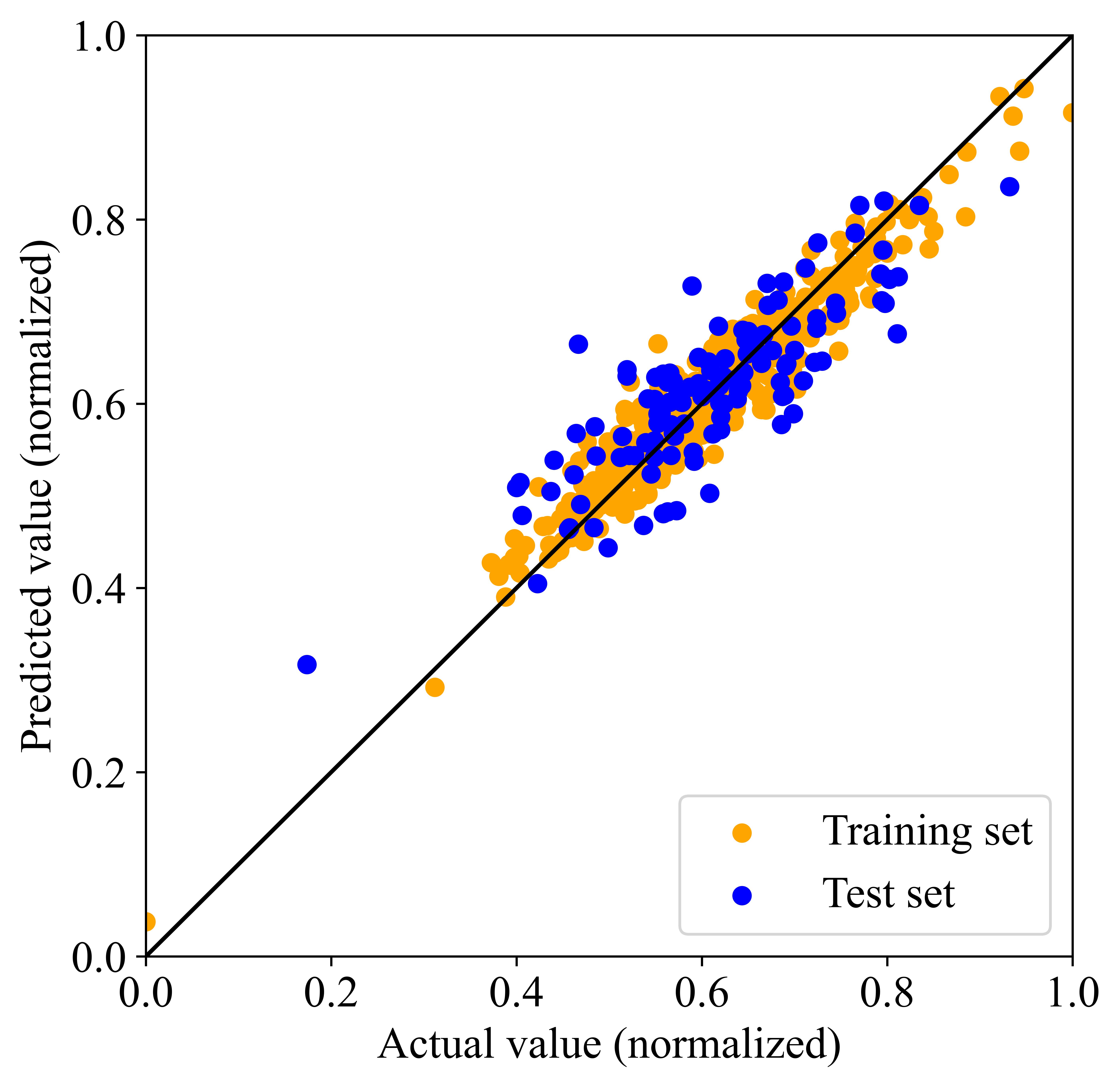}
\caption{Training set performance metrics: MSE: 0.067, MAE: 0.198, R\textsuperscript{2}: 0.914, MAPE: 1.26\%. Test set performance metrics: MSE: 0.264, MAE: 0.409, R\textsuperscript{2}: 0.722, MAPE: 2.62\%.}
\label{fig:regression-performance} 
\end{figure}

In the present work, regression is used as a tool that allows for the prediction of the average coating thickness for each production run, using fewer measurements than the 15 currently used. Specifically, we use the features identified through clustering and five thickness measurements (the closest to the reactor's inlet (R\textsubscript{0})) as inputs. This leads to accurate prediction of the mean coating thickness (average of R\textsubscript{1/2} and R) for both training ($\mathrm{R^2=0.914}$) and the test set ($\mathrm{R^2=0.722}$) (cf. \cref{fig:regression-performance}). This method proves valuable for streamlined post-production quality control as it allows for precise quality assessment with only one third of the previously required measurements.

\subsection{Shapley value analysis}\label{sec:shapley-analysis}
\begin{figure}[h!]
\captionsetup[subfigure]{justification=centering}
\centering
\begin{subfigure}[b]{0.49\textwidth}
   \includegraphics[width=1\textwidth]{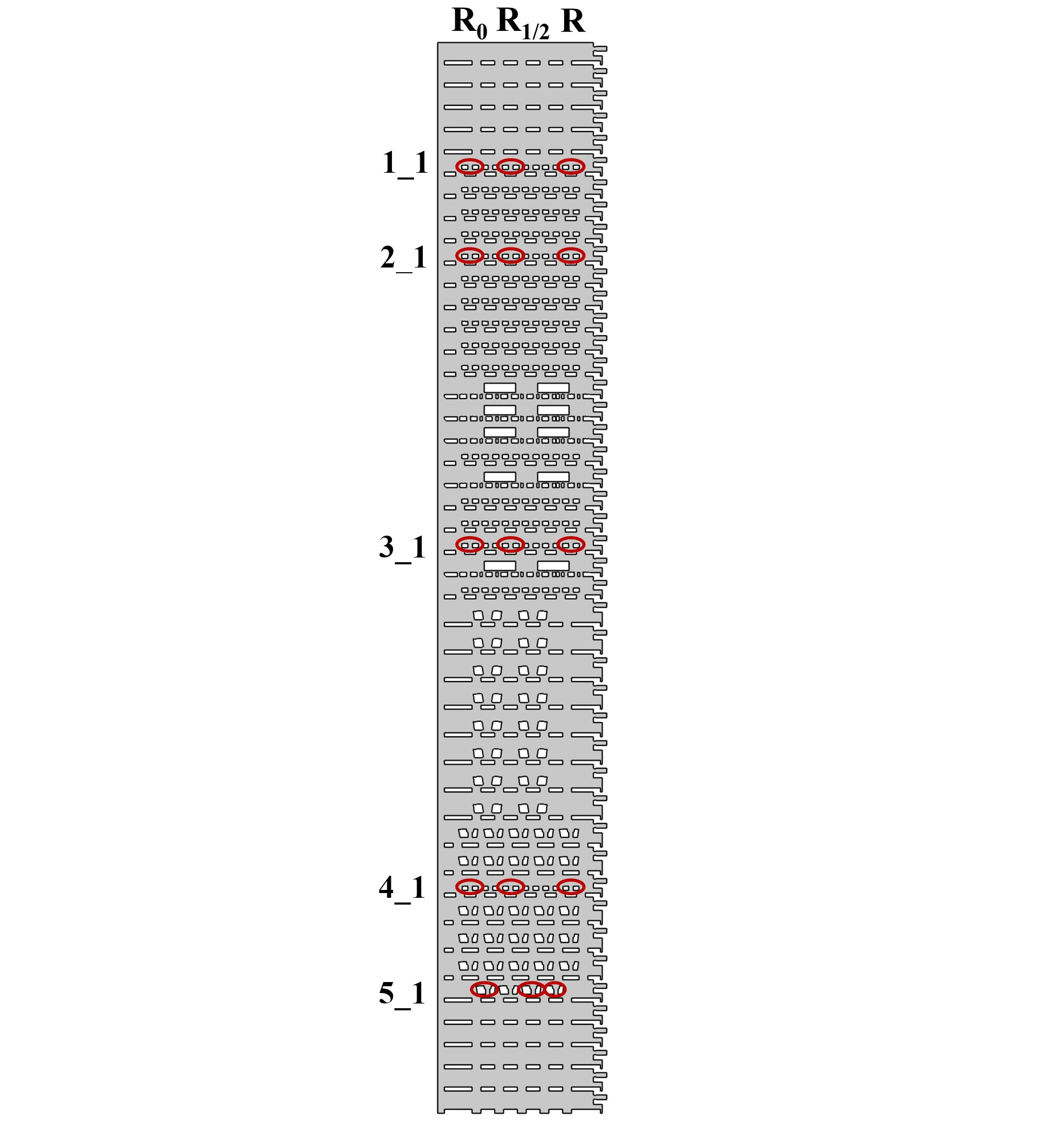}
   \caption{}
   \label{fig:3dgeopositions-annotated} 
\end{subfigure}
\hfill
\begin{subfigure}[b]{0.49\textwidth}
   \includegraphics[width=1\textwidth]{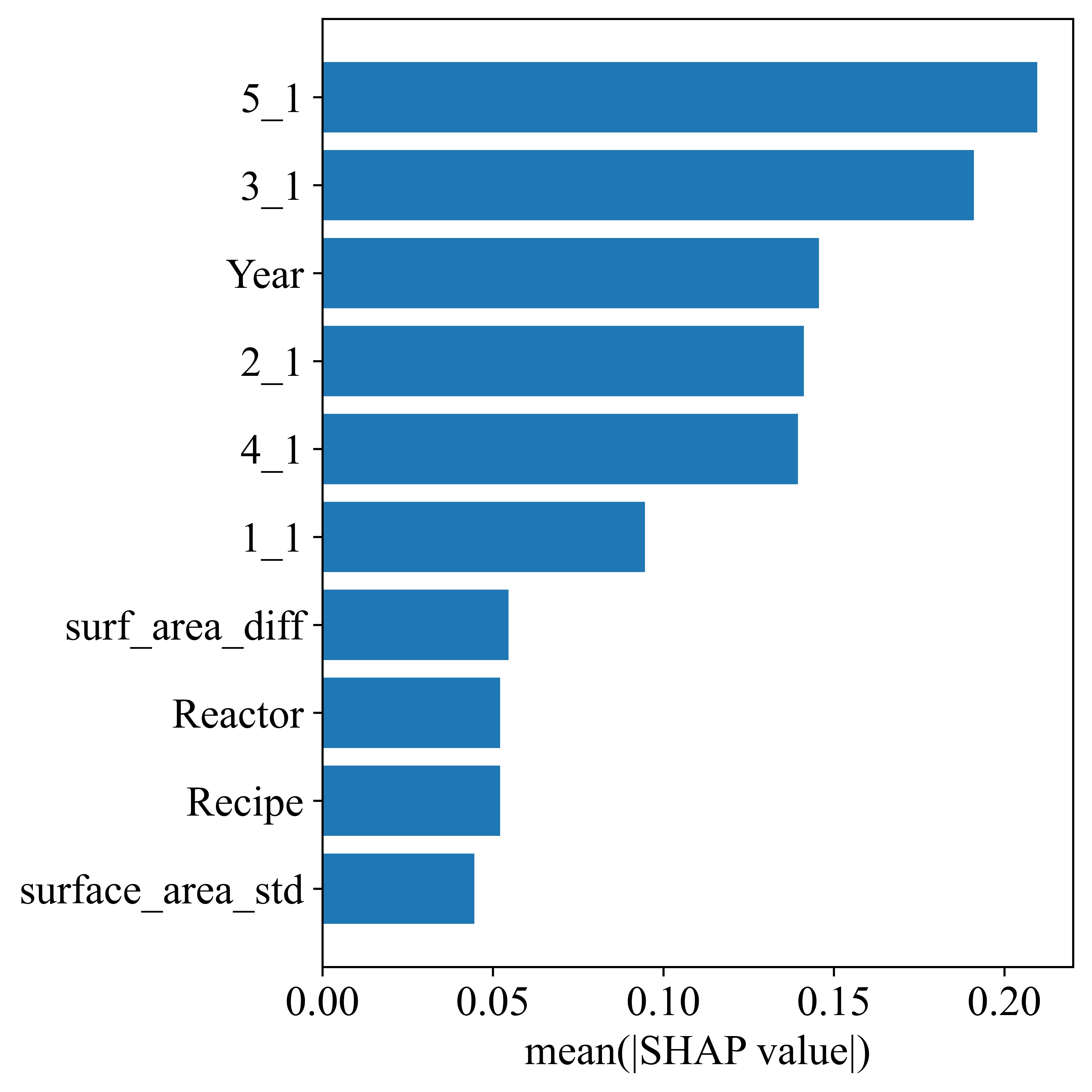}
   \caption{}
   \label{fig:shap-outer-mean}
\end{subfigure}
\caption{(a) 2D representation of the reactor, indicating the positions of the thickness measurements used as inputs for the regression problem. (b) Calculated mean absolute SHAP values for each of the inputs to the regression model. The five provided thickness measurements along with the year of production appear to be the dominant features, followed by surf_area_diff, the reactor and the recipe used for production and the standard deviation of surface area within the reactor.}
\label{fig:shap-n-annotated-3d}
\end{figure}

The most influential features that affect the predicted average coating thickness are identified by computing the SHAP values for the developed regression model. The mean absolute SHAP values are shown in \cref{fig:shap-outer-mean}. The five thickness measurements provided along with the year of production emerge as the most crucial features. Of the five thickness measurements provided, the lowest contribution comes from the measurement on the first disk from the top of the reactor. They are followed by the four remaining features, i.e., recipe, difference between the nominal and actual substrate surface area within the reactor (surf\_area\_diff), standard deviation of the surface area (surface\_area\_std) and the reactor used for production. These four features demonstrate a similar contribution to the model's predictions.

\section{Conclusions}
\label{sec:conclusion}
This study introduces a data-driven approach for uncovering patterns and influencing process inputs in an industrial Chemical Vapor Deposition (CVD) process, addressing challenges associated with process complexity and dataset characteristics.

Our analysis relies on subject matter expertise, combined with supervised and unsupervised learning methods. The main premise is that the performance of data-driven algorithms, given a specific dataset, is influenced by, and is indicative of, the importance of the inputs used during training. This is supported here by intuition about critical process inputs and \textit{some} knowledge about the important quality characteristics. 

We use unsupervised learning to obtain meaningful data labels that correspond to groups of production runs of similar quality. We then use these labels, in the context of supervised learning, to predict the outcome for a new set of inputs, thus providing a cost-efficient shortcut for quality control. 

The importance of features in investigated using Shapley values, which corroborates both subject matter expertise and also the conclusions drawn from the accuracy of classification methods. The results of this study offer opportunities to streamline post-production quality control and contribute to the ongoing refinement of the manufacturing process.

It is worth noting that this framework is adaptable to other processes, contingent on data availability. Even in cases with limited data, this approach unveils potential process-determining inputs, corroborating the insights of process experts in a purely data-driven manner.

Furthermore, consistent and improved data collection in the coming years will not only aid in validating and enhancing the developed predictive models, but also contribute to the continuous optimization of the overall process.

\section*{Acknowledgments}
This research was funded in part by the Luxembourg National Research Fund (FNR), grant reference [16758846]. For the purpose of open access, the authors have applied a Creative Commons Attribution 4.0 International (CC BY 4.0) license to any Author Accepted Manuscript version arising from this submission. PP gratefully acknowledges funding from the FSTM in the University of Luxembourg. The work of IGK has been partly supported by the US Department of Energy.


\bibliographystyle{elsarticle-num-names} 
\bibliography{mylib}

\begin{thebibliography}{73}
\expandafter\ifx\csname natexlab\endcsname\relax\def\natexlab#1{#1}\fi
\providecommand{\url}[1]{\texttt{#1}}
\providecommand{\href}[2]{#2}
\providecommand{\path}[1]{#1}
\providecommand{\DOIprefix}{doi:}
\providecommand{\ArXivprefix}{arXiv:}
\providecommand{\URLprefix}{URL: }
\providecommand{\Pubmedprefix}{pmid:}
\providecommand{\doi}[1]{\href{http://dx.doi.org/#1}{\path{#1}}}
\providecommand{\Pubmed}[1]{\href{pmid:#1}{\path{#1}}}
\providecommand{\bibinfo}[2]{#2}
\ifx\xfnm\relax \def\xfnm[#1]{\unskip,\space#1}\fi
\bibitem[{Cote et~al.(1999)Cote, Nguyen, Stamper, Armbrust, Tobben, Conti, and Lee}]{cotePlasmaassistedChemicalVapor1999}
\bibinfo{author}{D.~R. Cote}, \bibinfo{author}{S.~V. Nguyen}, \bibinfo{author}{A.~K. Stamper}, \bibinfo{author}{D.~S. Armbrust}, \bibinfo{author}{D.~Tobben}, \bibinfo{author}{R.~A. Conti}, \bibinfo{author}{G.~Y. Lee},
\newblock \bibinfo{title}{Plasma-assisted chemical vapor deposition of dielectric thin films for {{ULSI}} semiconductor circuits},
\newblock \bibinfo{journal}{IBM Journal of Research and Development} \bibinfo{volume}{43} (\bibinfo{year}{1999}) \bibinfo{pages}{5--38}. \DOIprefix\doi{10.1147/rd.431.0005}.
\bibitem[{Biefeld(2002)}]{biefeldMetalorganicChemicalVapor2002}
\bibinfo{author}{R.~M. Biefeld},
\newblock \bibinfo{title}{The metal-organic chemical vapor deposition and properties of {{III}}--{{V}} antimony-based semiconductor materials},
\newblock \bibinfo{journal}{Materials Science and Engineering: R: Reports} \bibinfo{volume}{36} (\bibinfo{year}{2002}) \bibinfo{pages}{105--142}. \DOIprefix\doi{10.1016/S0927-796X(02)00002-5}.
\bibitem[{Ha et~al.(1996)Ha, Nam, Lim, Oh, and Hong}]{haPropertiesTiO2Membranes1996}
\bibinfo{author}{H.~Y. Ha}, \bibinfo{author}{S.~W. Nam}, \bibinfo{author}{T.~H. Lim}, \bibinfo{author}{I.-H. Oh}, \bibinfo{author}{S.-A. Hong},
\newblock \bibinfo{title}{Properties of the {{TiO2}} membranes prepared by {{CVD}} of titanium tetraisopropoxide},
\newblock \bibinfo{journal}{Journal of Membrane Science} \bibinfo{volume}{111} (\bibinfo{year}{1996}) \bibinfo{pages}{81--92}. \DOIprefix\doi{10.1016/0376-7388(95)00278-2}.
\bibitem[{Khatib and Oyama(2013)}]{khatibSilicaMembranesHydrogen2013}
\bibinfo{author}{S.~J. Khatib}, \bibinfo{author}{S.~T. Oyama},
\newblock \bibinfo{title}{Silica membranes for hydrogen separation prepared by chemical vapor deposition ({{CVD}})},
\newblock \bibinfo{journal}{Separation and Purification Technology} \bibinfo{volume}{111} (\bibinfo{year}{2013}) \bibinfo{pages}{20--42}. \DOIprefix\doi{10.1016/j.seppur.2013.03.032}.
\bibitem[{Schmauder et~al.(2006)Schmauder, Nauenburg, Kruse, and Ickes}]{schmauderHardCoatingsPlasma2006}
\bibinfo{author}{T.~Schmauder}, \bibinfo{author}{K.~D. Nauenburg}, \bibinfo{author}{K.~Kruse}, \bibinfo{author}{G.~Ickes},
\newblock \bibinfo{title}{Hard coatings by plasma {{CVD}} on polycarbonate for automotive and optical applications},
\newblock \bibinfo{journal}{Thin Solid Films} \bibinfo{volume}{502} (\bibinfo{year}{2006}) \bibinfo{pages}{270--274}. \DOIprefix\doi{10.1016/j.tsf.2005.07.296}.
\bibitem[{Jia et~al.(2021)Jia, Chen, Zhang, Lin, Guo, Lu, Li, Zhai, Ai, and Lou}]{jiaCVDGrowthHighquality2021}
\bibinfo{author}{S.~Jia}, \bibinfo{author}{W.~Chen}, \bibinfo{author}{J.~Zhang}, \bibinfo{author}{C.~Y. Lin}, \bibinfo{author}{H.~Guo}, \bibinfo{author}{G.~Lu}, \bibinfo{author}{K.~Li}, \bibinfo{author}{T.~Zhai}, \bibinfo{author}{Q.~Ai}, \bibinfo{author}{J.~Lou},
\newblock \bibinfo{title}{{{CVD}} growth of high-quality and large-area continuous h-{{BN}} thin films directly on stainless-steel as protective coatings},
\newblock \bibinfo{journal}{Materials Today Nano} \bibinfo{volume}{16} (\bibinfo{year}{2021}) \bibinfo{pages}{100135}. \DOIprefix\doi{10.1016/j.mtnano.2021.100135}.
\bibitem[{Karner et~al.(1996)Karner, Pedrazzini, Reineck, Sj{\"o}strand, and Bergmann}]{karnerCVDDiamondCoated1996}
\bibinfo{author}{J.~Karner}, \bibinfo{author}{M.~Pedrazzini}, \bibinfo{author}{I.~Reineck}, \bibinfo{author}{M.~E. Sj{\"o}strand}, \bibinfo{author}{E.~Bergmann},
\newblock \bibinfo{title}{{{CVD}} diamond coated cemented carbide cutting tools},
\newblock \bibinfo{journal}{Materials Science and Engineering: A} \bibinfo{volume}{209} (\bibinfo{year}{1996}) \bibinfo{pages}{405--413}. \DOIprefix\doi{10.1016/0921-5093(95)10140-3}.
\bibitem[{Kathrein et~al.(2003)Kathrein, Schintlmeister, Wallgram, and Schleinkofer}]{kathreinDopedCVDAl2O32003}
\bibinfo{author}{M.~Kathrein}, \bibinfo{author}{W.~Schintlmeister}, \bibinfo{author}{W.~Wallgram}, \bibinfo{author}{U.~Schleinkofer},
\newblock \bibinfo{title}{Doped {{CVD Al2O3}} coatings for high performance cutting tools},
\newblock \bibinfo{journal}{Surface and Coatings Technology} \bibinfo{volume}{163--164} (\bibinfo{year}{2003}) \bibinfo{pages}{181--188}. \DOIprefix\doi{10.1016/s0257-8972(02)00483-8}.
\bibitem[{Mitrovic et~al.(2007)Mitrovic, Gurary, and Quinn}]{mitrovicProcessConditionsOptimization2007}
\bibinfo{author}{B.~Mitrovic}, \bibinfo{author}{A.~Gurary}, \bibinfo{author}{W.~Quinn},
\newblock \bibinfo{title}{Process conditions optimization for the maximum deposition rate and uniformity in vertical rotating disc {{MOCVD}} reactors based on {{CFD}} modeling},
\newblock \bibinfo{journal}{Journal of Crystal Growth} \bibinfo{volume}{303} (\bibinfo{year}{2007}) \bibinfo{pages}{323--329}. \DOIprefix\doi{10.1016/j.jcrysgro.2006.11.247}.
\bibitem[{Cheimarios et~al.(2012)Cheimarios, Koronaki, and Boudouvis}]{cheimariosIlluminatingNonlinearDependence2012}
\bibinfo{author}{N.~Cheimarios}, \bibinfo{author}{E.~D. Koronaki}, \bibinfo{author}{A.~G. Boudouvis},
\newblock \bibinfo{title}{Illuminating nonlinear dependence of film deposition rate in a {{CVD}} reactor on operating conditions},
\newblock \bibinfo{journal}{Chemical Engineering Journal} \bibinfo{volume}{181--182} (\bibinfo{year}{2012}) \bibinfo{pages}{516--523}. \DOIprefix\doi{10.1016/j.cej.2011.11.008}.
\bibitem[{Koronaki et~al.(2014)Koronaki, Cheimarios, Laux, and Boudouvis}]{koronakiNonAxisymmetricFlowFields2014}
\bibinfo{author}{E.~D. Koronaki}, \bibinfo{author}{N.~Cheimarios}, \bibinfo{author}{H.~Laux}, \bibinfo{author}{A.~G. Boudouvis},
\newblock \bibinfo{title}{Non-{{Axisymmetric Flow Fields}} in {{Axisymmetric CVD Reactor Setups Revisited}}: {{Influence}} on the {{Film}}'s {{Non-Uniformity}}},
\newblock \bibinfo{journal}{ECS Solid State Lett.} \bibinfo{volume}{3} (\bibinfo{year}{2014}) \bibinfo{pages}{P37}. \DOIprefix\doi{10.1149/2.002404ssl}.
\bibitem[{Gakis et~al.(2015)Gakis, Koronaki, and Boudouvis}]{gakisNumericalInvestigationMultiple2015}
\bibinfo{author}{G.~Gakis}, \bibinfo{author}{E.~Koronaki}, \bibinfo{author}{A.~Boudouvis},
\newblock \bibinfo{title}{Numerical investigation of multiple stationary and time-periodic flow regimes in vertical rotating disc {{CVD}} reactors},
\newblock \bibinfo{journal}{Journal of Crystal Growth} \bibinfo{volume}{432} (\bibinfo{year}{2015}) \bibinfo{pages}{152--159}. \DOIprefix\doi{10.1016/j.jcrysgro.2015.09.026}.
\bibitem[{Koronaki et~al.(2016)Koronaki, Gakis, Cheimarios, and Boudouvis}]{koronakiEfficientTracingStability2016}
\bibinfo{author}{E.~D. Koronaki}, \bibinfo{author}{G.~P. Gakis}, \bibinfo{author}{N.~Cheimarios}, \bibinfo{author}{A.~G. Boudouvis},
\newblock \bibinfo{title}{Efficient tracing and stability analysis of multiple stationary and periodic states with exploitation of commercial {{CFD}} software},
\newblock \bibinfo{journal}{Chemical Engineering Science} \bibinfo{volume}{150} (\bibinfo{year}{2016}) \bibinfo{pages}{26--34}. \DOIprefix\doi{10.1016/j.ces.2016.04.043}.
\bibitem[{Aviziotis et~al.(2016)Aviziotis, Cheimarios, Duguet, Vahlas, and Boudouvis}]{aviziotisMultiscaleModelingExperimental2016}
\bibinfo{author}{I.~G. Aviziotis}, \bibinfo{author}{N.~Cheimarios}, \bibinfo{author}{T.~Duguet}, \bibinfo{author}{C.~Vahlas}, \bibinfo{author}{A.~G. Boudouvis},
\newblock \bibinfo{title}{Multiscale modeling and experimental analysis of chemical vapor deposited aluminum films: {{Linking}} reactor operating conditions with roughness evolution},
\newblock \bibinfo{journal}{Chemical Engineering Science} \bibinfo{volume}{155} (\bibinfo{year}{2016}) \bibinfo{pages}{449--458}. \DOIprefix\doi{10.1016/j.ces.2016.08.039}.
\bibitem[{Aviziotis et~al.(2017)Aviziotis, Duguet, Vahlas, and Boudouvis}]{aviziotisCombinedMacroNanoscale2017}
\bibinfo{author}{I.~G. Aviziotis}, \bibinfo{author}{T.~Duguet}, \bibinfo{author}{C.~Vahlas}, \bibinfo{author}{A.~G. Boudouvis},
\newblock \bibinfo{title}{Combined {{Macro}}/{{Nanoscale Investigation}} of the {{Chemical Vapor Deposition}} of {{Fe}} from {{Fe}}({{CO}})5},
\newblock \bibinfo{journal}{Advanced Materials Interfaces} \bibinfo{volume}{4} (\bibinfo{year}{2017}) \bibinfo{pages}{1601185}. \DOIprefix\doi{10.1002/admi.201601185}.
\bibitem[{Psarellis et~al.(2018)Psarellis, Aviziotis, Duguet, Vahlas, Koronaki, and Boudouvis}]{psarellisInvestigationReactionMechanisms2018}
\bibinfo{author}{G.~M. Psarellis}, \bibinfo{author}{I.~G. Aviziotis}, \bibinfo{author}{T.~Duguet}, \bibinfo{author}{C.~Vahlas}, \bibinfo{author}{E.~D. Koronaki}, \bibinfo{author}{A.~G. Boudouvis},
\newblock \bibinfo{title}{Investigation of reaction mechanisms in the chemical vapor deposition of al from {{DMEAA}}},
\newblock \bibinfo{journal}{Chemical Engineering Science} \bibinfo{volume}{177} (\bibinfo{year}{2018}) \bibinfo{pages}{464--470}. \DOIprefix\doi{10.1016/j.ces.2017.12.006}.
\bibitem[{Topka et~al.(2022)Topka, Vergnes, Tsiros, Papavasileiou, Decosterd, Diallo, Senocq, Samelor, Pellerin, Menu, Vahlas, and Caussat}]{topkaInnovativeKineticModel2022}
\bibinfo{author}{K.~C. Topka}, \bibinfo{author}{H.~Vergnes}, \bibinfo{author}{T.~Tsiros}, \bibinfo{author}{P.~Papavasileiou}, \bibinfo{author}{L.~Decosterd}, \bibinfo{author}{B.~Diallo}, \bibinfo{author}{F.~Senocq}, \bibinfo{author}{D.~Samelor}, \bibinfo{author}{N.~Pellerin}, \bibinfo{author}{M.-J. Menu}, \bibinfo{author}{C.~Vahlas}, \bibinfo{author}{B.~Caussat},
\newblock \bibinfo{title}{An innovative kinetic model allowing insight in the moderate temperature chemical vapor deposition of silicon oxynitride films from tris(dimethylsilyl)amine},
\newblock \bibinfo{journal}{Chemical Engineering Journal} \bibinfo{volume}{431} (\bibinfo{year}{2022}) \bibinfo{pages}{133350}. \DOIprefix\doi{10.1016/j.cej.2021.133350}.
\bibitem[{Gkinis et~al.(2017)Gkinis, Aviziotis, Koronaki, Gakis, and Boudouvis}]{gkinisEffectsFlowMultiplicity2017}
\bibinfo{author}{P.~Gkinis}, \bibinfo{author}{I.~Aviziotis}, \bibinfo{author}{E.~Koronaki}, \bibinfo{author}{G.~Gakis}, \bibinfo{author}{A.~Boudouvis},
\newblock \bibinfo{title}{The effects of flow multiplicity on {{GaN}} deposition in a rotating disk {{CVD}} reactor},
\newblock \bibinfo{journal}{Journal of Crystal Growth} \bibinfo{volume}{458} (\bibinfo{year}{2017}) \bibinfo{pages}{140--148}. \DOIprefix\doi{10.1016/j.jcrysgro.2016.10.065}.
\bibitem[{Koronaki et~al.(2019)Koronaki, Gkinis, Beex, Bordas, Theodoropoulos, and Boudouvis}]{koronakiClassificationStatesModel2019}
\bibinfo{author}{E.~Koronaki}, \bibinfo{author}{P.~Gkinis}, \bibinfo{author}{L.~Beex}, \bibinfo{author}{S.~Bordas}, \bibinfo{author}{C.~Theodoropoulos}, \bibinfo{author}{A.~Boudouvis},
\newblock \bibinfo{title}{Classification of states and model order reduction of large scale {{Chemical Vapor Deposition}} processes with solution multiplicity},
\newblock \bibinfo{journal}{Computers \& Chemical Engineering} \bibinfo{volume}{121} (\bibinfo{year}{2019}) \bibinfo{pages}{148--157}. \DOIprefix\doi{10.1016/j.compchemeng.2018.08.023}.
\bibitem[{Saxena and Saad(2007)}]{saxenaEvolvingArtificialNeural2007}
\bibinfo{author}{A.~Saxena}, \bibinfo{author}{A.~Saad},
\newblock \bibinfo{title}{Evolving an artificial neural network classifier for condition monitoring of rotating mechanical systems},
\newblock \bibinfo{journal}{Applied Soft Computing} \bibinfo{volume}{7} (\bibinfo{year}{2007}) \bibinfo{pages}{441--454}. \DOIprefix\doi{10.1016/j.asoc.2005.10.001}.
\bibitem[{Susto et~al.(2015)Susto, Schirru, Pampuri, McLoone, and Beghi}]{sustoMachineLearningPredictive2015}
\bibinfo{author}{G.~A. Susto}, \bibinfo{author}{A.~Schirru}, \bibinfo{author}{S.~Pampuri}, \bibinfo{author}{S.~McLoone}, \bibinfo{author}{A.~Beghi},
\newblock \bibinfo{title}{Machine {{Learning}} for {{Predictive Maintenance}}: {{A Multiple Classifier Approach}}},
\newblock \bibinfo{journal}{IEEE Trans. Ind. Inf.} \bibinfo{volume}{11} (\bibinfo{year}{2015}) \bibinfo{pages}{812--820}. \DOIprefix\doi{10.1109/tii.2014.2349359}.
\bibitem[{Wu et~al.(2019)Wu, Yu, and Wang}]{wuExperimentalStudyProcess2019}
\bibinfo{author}{H.~Wu}, \bibinfo{author}{Z.~Yu}, \bibinfo{author}{Y.~Wang},
\newblock \bibinfo{title}{Experimental study of the process failure diagnosis in additive manufacturing based on acoustic emission},
\newblock \bibinfo{journal}{Measurement} \bibinfo{volume}{136} (\bibinfo{year}{2019}) \bibinfo{pages}{445--453}. \DOIprefix\doi{10.1016/j.measurement.2018.12.067}.
\bibitem[{Priore et~al.(2018)Priore, Ponte, Puente, and G{\'o}mez}]{prioreLearningbasedSchedulingFlexible2018}
\bibinfo{author}{P.~Priore}, \bibinfo{author}{B.~Ponte}, \bibinfo{author}{J.~Puente}, \bibinfo{author}{A.~G{\'o}mez},
\newblock \bibinfo{title}{Learning-based scheduling of flexible manufacturing systems using ensemble methods},
\newblock \bibinfo{journal}{Computers \& Industrial Engineering} \bibinfo{volume}{126} (\bibinfo{year}{2018}) \bibinfo{pages}{282--291}. \DOIprefix\doi{10.1016/j.cie.2018.09.034}.
\bibitem[{Agarwal et~al.(2020)Agarwal, Tamer, Sahraei, and Budman}]{agarwalDeepLearningClassification2020}
\bibinfo{author}{P.~Agarwal}, \bibinfo{author}{M.~Tamer}, \bibinfo{author}{M.~H. Sahraei}, \bibinfo{author}{H.~Budman},
\newblock \bibinfo{title}{Deep {{Learning}} for {{Classification}} of {{Profit-Based Operating Regions}} in {{Industrial Processes}}},
\newblock \bibinfo{journal}{Ind. Eng. Chem. Res.} \bibinfo{volume}{59} (\bibinfo{year}{2020}) \bibinfo{pages}{2378--2395}. \DOIprefix\doi{10.1021/acs.iecr.9b04737}.
\bibitem[{Papananias et~al.(2019)Papananias, McLeay, Mahfouf, and Kadirkamanathan}]{papananiasBayesianFrameworkEstimate2019}
\bibinfo{author}{M.~Papananias}, \bibinfo{author}{T.~E. McLeay}, \bibinfo{author}{M.~Mahfouf}, \bibinfo{author}{V.~Kadirkamanathan},
\newblock \bibinfo{title}{A {{Bayesian}} framework to estimate part quality and associated uncertainties in multistage manufacturing},
\newblock \bibinfo{journal}{Computers in Industry} \bibinfo{volume}{105} (\bibinfo{year}{2019}) \bibinfo{pages}{35--47}. \DOIprefix\doi{10.1016/j.compind.2018.10.008}.
\bibitem[{Papavasileiou et~al.(2023)Papavasileiou, Koronaki, Pozzetti, Kathrein, Czettl, Boudouvis, and Bordas}]{papavasileiouEquationbasedDatadrivenModeling2023}
\bibinfo{author}{P.~Papavasileiou}, \bibinfo{author}{E.~D. Koronaki}, \bibinfo{author}{G.~Pozzetti}, \bibinfo{author}{M.~Kathrein}, \bibinfo{author}{C.~Czettl}, \bibinfo{author}{A.~G. Boudouvis}, \bibinfo{author}{S.~P. Bordas},
\newblock \bibinfo{title}{Equation-based and data-driven modeling strategies for industrial coating processes},
\newblock \bibinfo{journal}{Computers in Industry} \bibinfo{volume}{149} (\bibinfo{year}{2023}) \bibinfo{pages}{103938}. \DOIprefix\doi{10.1016/j.compind.2023.103938}.
\bibitem[{Ma et~al.(2019)Ma, Zhu, Benton, and Romagnoli}]{maContinuousControlPolymerization2019}
\bibinfo{author}{Y.~Ma}, \bibinfo{author}{W.~Zhu}, \bibinfo{author}{M.~G. Benton}, \bibinfo{author}{J.~Romagnoli},
\newblock \bibinfo{title}{Continuous control of a polymerization system with deep reinforcement learning},
\newblock \bibinfo{journal}{Journal of Process Control} \bibinfo{volume}{75} (\bibinfo{year}{2019}) \bibinfo{pages}{40--47}. \DOIprefix\doi{10.1016/j.jprocont.2018.11.004}.
\bibitem[{Humfeld et~al.(2021)Humfeld, Gu, Butler, Nelson, and Zobeiry}]{humfeldMachineLearningFramework2021}
\bibinfo{author}{K.~D. Humfeld}, \bibinfo{author}{D.~Gu}, \bibinfo{author}{G.~A. Butler}, \bibinfo{author}{K.~Nelson}, \bibinfo{author}{N.~Zobeiry},
\newblock \bibinfo{title}{A machine learning framework for real-time inverse modeling and multi-objective process optimization of composites for active manufacturing control},
\newblock \bibinfo{journal}{Composites Part B: Engineering} \bibinfo{volume}{223} (\bibinfo{year}{2021}) \bibinfo{pages}{109150}. \DOIprefix\doi{10.1016/j.compositesb.2021.109150}.
\bibitem[{Gkinis et~al.(2019)Gkinis, Koronaki, Skouteris, Aviziotis, and Boudouvis}]{gkinisBuildingDatadrivenReduced2019}
\bibinfo{author}{P.~Gkinis}, \bibinfo{author}{E.~Koronaki}, \bibinfo{author}{A.~Skouteris}, \bibinfo{author}{I.~Aviziotis}, \bibinfo{author}{A.~Boudouvis},
\newblock \bibinfo{title}{Building a data-driven reduced order model of a chemical vapor deposition process from low-fidelity {{CFD}} simulations},
\newblock \bibinfo{journal}{Chemical Engineering Science} \bibinfo{volume}{199} (\bibinfo{year}{2019}) \bibinfo{pages}{371--380}. \DOIprefix\doi{10.1016/j.ces.2019.01.009}.
\bibitem[{Spencer et~al.(2021)Spencer, Gkinis, Koronaki, Gerogiorgis, Bordas, and Boudouvis}]{spencerInvestigationChemicalVapor2021}
\bibinfo{author}{R.~Spencer}, \bibinfo{author}{P.~Gkinis}, \bibinfo{author}{E.~Koronaki}, \bibinfo{author}{D.~Gerogiorgis}, \bibinfo{author}{S.~Bordas}, \bibinfo{author}{A.~Boudouvis},
\newblock \bibinfo{title}{Investigation of the chemical vapor deposition of {{Cu}} from copper amidinate through data driven efficient {{CFD}} modelling},
\newblock \bibinfo{journal}{Computers \& Chemical Engineering} \bibinfo{volume}{149} (\bibinfo{year}{2021}) \bibinfo{pages}{107289}. \DOIprefix\doi{10.1016/j.compchemeng.2021.107289}.
\bibitem[{{Martin-Linares} et~al.(2023){Martin-Linares}, Psarellis, Karapetsas, Koronaki, and Kevrekidis}]{martin-linaresPhysicsagnosticPhysicsinfusedMachine2023}
\bibinfo{author}{C.~P. {Martin-Linares}}, \bibinfo{author}{Y.~M. Psarellis}, \bibinfo{author}{G.~Karapetsas}, \bibinfo{author}{E.~D. Koronaki}, \bibinfo{author}{I.~G. Kevrekidis},
\newblock \bibinfo{title}{Physics-agnostic and physics-infused machine learning for thin films flows: Modelling, and predictions from small data},
\newblock \bibinfo{journal}{Journal of Fluid Mechanics} \bibinfo{volume}{975} (\bibinfo{year}{2023}) \bibinfo{pages}{A41}. \DOIprefix\doi{10.1017/jfm.2023.868}.
\bibitem[{Shapley(1952)}]{shapleyValueNPersonGames1952}
\bibinfo{author}{L.~S. Shapley}, \bibinfo{title}{A {{Value}} for {{N-Person Games}}}, \bibinfo{type}{Technical Report}, RAND Corporation, \bibinfo{year}{1952}.
\bibitem[{Lundberg and Lee(2017)}]{lundbergUnifiedApproachInterpreting2017}
\bibinfo{author}{S.~M. Lundberg}, \bibinfo{author}{S.-I. Lee},
\newblock \bibinfo{title}{A {{Unified Approach}} to {{Interpreting Model Predictions}}},
\newblock in: \bibinfo{booktitle}{Advances in {{Neural Information Processing Systems}}}, volume~\bibinfo{volume}{30}, \bibinfo{publisher}{Curran Associates, Inc.}, \bibinfo{year}{2017}.
\bibitem[{Barredo~Arrieta et~al.(2020)Barredo~Arrieta, {D{\'i}az-Rodr{\'i}guez}, Del~Ser, Bennetot, Tabik, Barbado, Garcia, {Gil-Lopez}, Molina, Benjamins, Chatila, and Herrera}]{barredoarrietaExplainableArtificialIntelligence2020a}
\bibinfo{author}{A.~Barredo~Arrieta}, \bibinfo{author}{N.~{D{\'i}az-Rodr{\'i}guez}}, \bibinfo{author}{J.~Del~Ser}, \bibinfo{author}{A.~Bennetot}, \bibinfo{author}{S.~Tabik}, \bibinfo{author}{A.~Barbado}, \bibinfo{author}{S.~Garcia}, \bibinfo{author}{S.~{Gil-Lopez}}, \bibinfo{author}{D.~Molina}, \bibinfo{author}{R.~Benjamins}, \bibinfo{author}{R.~Chatila}, \bibinfo{author}{F.~Herrera},
\newblock \bibinfo{title}{Explainable {{Artificial Intelligence}} ({{XAI}}): {{Concepts}}, taxonomies, opportunities and challenges toward responsible {{AI}}},
\newblock \bibinfo{journal}{Information Fusion} \bibinfo{volume}{58} (\bibinfo{year}{2020}) \bibinfo{pages}{82--115}. \DOIprefix\doi{10.1016/j.inffus.2019.12.012}.
\bibitem[{Sundararajan and Najmi(2020)}]{sundararajanManyShapleyValues2020}
\bibinfo{author}{M.~Sundararajan}, \bibinfo{author}{A.~Najmi},
\newblock \bibinfo{title}{The {{Many Shapley Values}} for {{Model Explanation}}},
\newblock in: \bibinfo{booktitle}{Proceedings of the 37th {{International Conference}} on {{Machine Learning}}}, \bibinfo{publisher}{PMLR}, \bibinfo{year}{2020}, pp. \bibinfo{pages}{9269--9278}.
\bibitem[{Chong and Jun(2005)}]{chongPerformanceVariableSelection2005}
\bibinfo{author}{I.-G. Chong}, \bibinfo{author}{C.-H. Jun},
\newblock \bibinfo{title}{Performance of some variable selection methods when multicollinearity is present},
\newblock \bibinfo{journal}{Chemometrics and Intelligent Laboratory Systems} \bibinfo{volume}{78} (\bibinfo{year}{2005}) \bibinfo{pages}{103--112}. \DOIprefix\doi{10.1016/j.chemolab.2004.12.011}.
\bibitem[{Lu et~al.(2014)Lu, Castillo, Chiang, and Edgar}]{luIndustrialPLSModel2014}
\bibinfo{author}{B.~Lu}, \bibinfo{author}{I.~Castillo}, \bibinfo{author}{L.~Chiang}, \bibinfo{author}{T.~F. Edgar},
\newblock \bibinfo{title}{Industrial {{PLS}} model variable selection using moving window variable importance in projection},
\newblock \bibinfo{journal}{Chemometrics and Intelligent Laboratory Systems} \bibinfo{volume}{135} (\bibinfo{year}{2014}) \bibinfo{pages}{90--109}. \DOIprefix\doi{10.1016/j.chemolab.2014.03.020}.
\bibitem[{Garthwaite(1994)}]{garthwaiteInterpretationPartialLeast1994}
\bibinfo{author}{P.~H. Garthwaite},
\newblock \bibinfo{title}{An {{Interpretation}} of {{Partial Least Squares}}},
\newblock \bibinfo{journal}{Journal of the American Statistical Association} \bibinfo{volume}{89} (\bibinfo{year}{1994}) \bibinfo{pages}{122--127}. \DOIprefix\doi{10.1080/01621459.1994.10476452}.
\bibitem[{Kumar(2021)}]{kumarPartialLeastSquare2021}
\bibinfo{author}{K.~Kumar},
\newblock \bibinfo{title}{Partial {{Least Square}} ({{PLS}}) {{Analysis}}: {{Most Favorite Tool}} in {{Chemometrics}} to {{Build}} a {{Calibration Model}}},
\newblock \bibinfo{journal}{Reson} \bibinfo{volume}{26} (\bibinfo{year}{2021}) \bibinfo{pages}{429--442}. \DOIprefix\doi{10.1007/s12045-021-1140-1}.
\bibitem[{Heinze et~al.(2018)Heinze, Wallisch, and Dunkler}]{heinzeVariableSelectionReview2018}
\bibinfo{author}{G.~Heinze}, \bibinfo{author}{C.~Wallisch}, \bibinfo{author}{D.~Dunkler},
\newblock \bibinfo{title}{Variable selection -- {{A}} review and recommendations for the practicing statistician},
\newblock \bibinfo{journal}{Biometrical Journal} \bibinfo{volume}{60} (\bibinfo{year}{2018}) \bibinfo{pages}{431--449}. \DOIprefix\doi{10.1002/bimj.201700067}.
\bibitem[{Koronaki et~al.(2020)Koronaki, Nikas, and Boudouvis}]{koronakiDatadrivenReducedorderModel2020}
\bibinfo{author}{E.~D. Koronaki}, \bibinfo{author}{A.~M. Nikas}, \bibinfo{author}{A.~G. Boudouvis},
\newblock \bibinfo{title}{A data-driven reduced-order model of nonlinear processes based on diffusion maps and artificial neural networks},
\newblock \bibinfo{journal}{Chemical Engineering Journal} \bibinfo{volume}{397} (\bibinfo{year}{2020}) \bibinfo{pages}{125475}. \DOIprefix\doi{10.1016/j.cej.2020.125475}.
\bibitem[{Koronaki et~al.(2023)Koronaki, Evangelou, Psarellis, Boudouvis, and Kevrekidis}]{koronakiPartialDataOutofsample2023a}
\bibinfo{author}{E.~D. Koronaki}, \bibinfo{author}{N.~Evangelou}, \bibinfo{author}{Y.~M. Psarellis}, \bibinfo{author}{A.~G. Boudouvis}, \bibinfo{author}{I.~G. Kevrekidis},
\newblock \bibinfo{title}{From partial data to out-of-sample parameter and observation estimation with diffusion maps and geometric harmonics},
\newblock \bibinfo{journal}{Computers \& Chemical Engineering}  (\bibinfo{year}{2023}) \bibinfo{pages}{108357}. \DOIprefix\doi{10.1016/j.compchemeng.2023.108357}.
\bibitem[{Brouwer and Eisenberg(2018)}]{brouwerUnderlyingConnectionsIdentifiability2018}
\bibinfo{author}{A.~F. Brouwer}, \bibinfo{author}{M.~C. Eisenberg}, \bibinfo{title}{The underlying connections between identifiability, active subspaces, and parameter space dimension reduction}, \bibinfo{year}{2018}. \DOIprefix\doi{10.48550/arXiv.1802.05641}. \href{http://arxiv.org/abs/1802.05641}{{\tt arXiv:1802.05641}}.
\bibitem[{Evangelou et~al.(2022)Evangelou, Wichrowski, Kevrekidis, Dietrich, Kooshkbaghi, McFann, and Kevrekidis}]{evangelouParameterCombinationsThat2022}
\bibinfo{author}{N.~Evangelou}, \bibinfo{author}{N.~J. Wichrowski}, \bibinfo{author}{G.~A. Kevrekidis}, \bibinfo{author}{F.~Dietrich}, \bibinfo{author}{M.~Kooshkbaghi}, \bibinfo{author}{S.~McFann}, \bibinfo{author}{I.~G. Kevrekidis},
\newblock \bibinfo{title}{On the parameter combinations that matter and on those that do not: Data-driven studies of parameter (non)identifiability},
\newblock \bibinfo{journal}{PNAS Nexus} \bibinfo{volume}{1} (\bibinfo{year}{2022}) \bibinfo{pages}{pgac154}. \DOIprefix\doi{10.1093/pnasnexus/pgac154}.
\bibitem[{Hochauer et~al.(2012)Hochauer, Mitterer, Penoy, Puchner, Michotte, Martinz, Hutter, and Kathrein}]{hochauerCarbonDopedAAl2O32012}
\bibinfo{author}{D.~Hochauer}, \bibinfo{author}{C.~Mitterer}, \bibinfo{author}{M.~Penoy}, \bibinfo{author}{S.~Puchner}, \bibinfo{author}{C.~Michotte}, \bibinfo{author}{H.~Martinz}, \bibinfo{author}{H.~Hutter}, \bibinfo{author}{M.~Kathrein},
\newblock \bibinfo{title}{Carbon doped {$\alpha$}-{{Al2O3}} coatings grown by chemical vapor deposition},
\newblock \bibinfo{journal}{Surface and Coatings Technology} \bibinfo{volume}{206} (\bibinfo{year}{2012}) \bibinfo{pages}{4771--4777}. \DOIprefix\doi{10.1016/j.surfcoat.2012.03.059}.
\bibitem[{{Bar-Hen} and Etsion(2017)}]{bar-henExperimentalStudyEffect2017}
\bibinfo{author}{M.~{Bar-Hen}}, \bibinfo{author}{I.~Etsion},
\newblock \bibinfo{title}{Experimental study of the effect of coating thickness and substrate roughness on tool wear during turning},
\newblock \bibinfo{journal}{Tribology International} \bibinfo{volume}{110} (\bibinfo{year}{2017}) \bibinfo{pages}{341--347}. \DOIprefix\doi{10.1016/j.triboint.2016.11.011}.
\bibitem[{{\L}{\k e}picka and {Gr{\k a}dzka-Dahlke}(2019)}]{lepickaInitialEvaluationPerformance2019}
\bibinfo{author}{M.~{\L}{\k e}picka}, \bibinfo{author}{M.~{Gr{\k a}dzka-Dahlke}},
\newblock \bibinfo{title}{The initial evaluation of performance of hard anti-wear coatings deposited on metallic substrates: Thickness, mechanical properties and adhesion measurements -- a brief review},
\newblock \bibinfo{journal}{REVIEWS ON ADVANCED MATERIALS SCIENCE} \bibinfo{volume}{58} (\bibinfo{year}{2019}) \bibinfo{pages}{50--65}. \DOIprefix\doi{10.1515/rams-2019-0003}.
\bibitem[{Papavasileiou et~al.(2022)Papavasileiou, Koronaki, Pozzetti, Kathrein, Czettl, Boudouvis, Mountziaris, and Bordas}]{papavasileiouEfficientChemistryenhancedCFD2022}
\bibinfo{author}{P.~Papavasileiou}, \bibinfo{author}{E.~D. Koronaki}, \bibinfo{author}{G.~Pozzetti}, \bibinfo{author}{M.~Kathrein}, \bibinfo{author}{C.~Czettl}, \bibinfo{author}{A.~G. Boudouvis}, \bibinfo{author}{T.~J. Mountziaris}, \bibinfo{author}{S.~P.~A. Bordas},
\newblock \bibinfo{title}{An efficient chemistry-enhanced {{CFD}} model for the investigation of the rate-limiting mechanisms in industrial {{Chemical Vapor Deposition}} reactors},
\newblock \bibinfo{journal}{Chemical Engineering Research and Design} \bibinfo{volume}{186} (\bibinfo{year}{2022}) \bibinfo{pages}{314--325}. \DOIprefix\doi{10.1016/j.cherd.2022.08.005}.
\bibitem[{Hastie et~al.(2009)Hastie, Tibshirani, and Friedman}]{hastieUnsupervisedLearning2009}
\bibinfo{author}{T.~Hastie}, \bibinfo{author}{R.~Tibshirani}, \bibinfo{author}{J.~Friedman},
\newblock \bibinfo{title}{Unsupervised {{Learning}}},
\newblock in: \bibinfo{editor}{T.~Hastie}, \bibinfo{editor}{R.~Tibshirani}, \bibinfo{editor}{J.~Friedman} (Eds.), \bibinfo{booktitle}{The {{Elements}} of {{Statistical Learning}}: {{Data Mining}}, {{Inference}}, and {{Prediction}}}, Springer {{Series}} in {{Statistics}}, \bibinfo{publisher}{Springer}, \bibinfo{address}{New York, NY}, \bibinfo{year}{2009}, pp. \bibinfo{pages}{485--585}. \DOIprefix\doi{10.1007/978-0-387-84858-7_14}.
\bibitem[{Wang et~al.(2016)Wang, Yao, and Zhao}]{wangAutoencoderBasedDimensionality2016}
\bibinfo{author}{Y.~Wang}, \bibinfo{author}{H.~Yao}, \bibinfo{author}{S.~Zhao},
\newblock \bibinfo{title}{Auto-encoder based dimensionality reduction},
\newblock \bibinfo{journal}{Neurocomputing} \bibinfo{volume}{184} (\bibinfo{year}{2016}) \bibinfo{pages}{232--242}. \DOIprefix\doi{10.1016/j.neucom.2015.08.104}.
\bibitem[{James et~al.(2021)James, Witten, Hastie, and Tibshirani}]{jamesUnsupervisedLearning2021}
\bibinfo{author}{G.~James}, \bibinfo{author}{D.~Witten}, \bibinfo{author}{T.~Hastie}, \bibinfo{author}{R.~Tibshirani},
\newblock \bibinfo{title}{Unsupervised {{Learning}}},
\newblock in: \bibinfo{editor}{G.~James}, \bibinfo{editor}{D.~Witten}, \bibinfo{editor}{T.~Hastie}, \bibinfo{editor}{R.~Tibshirani} (Eds.), \bibinfo{booktitle}{An {{Introduction}} to {{Statistical Learning}}: With {{Applications}} in {{R}}}, Springer {{Texts}} in {{Statistics}}, \bibinfo{publisher}{Springer US}, \bibinfo{address}{New York, NY}, \bibinfo{year}{2021}, pp. \bibinfo{pages}{497--552}. \DOIprefix\doi{10.1007/978-1-0716-1418-1_12}.
\bibitem[{MacQueen(1967)}]{macqueenMethodsClassificationAnalysis1967}
\bibinfo{author}{J.~MacQueen},
\newblock \bibinfo{title}{Some methods for classification and analysis of multivariate observations},
\newblock in: \bibinfo{booktitle}{Proceedings of the Fifth {{Berkeley}} Symposium on Mathematical Statistics and Probability}, volume~\bibinfo{volume}{1}, \bibinfo{publisher}{Oakland, CA, USA}, \bibinfo{year}{1967}, pp. \bibinfo{pages}{281--297}.
\bibitem[{Ankerst et~al.(1999)Ankerst, Breunig, Kriegel, and Sander}]{ankerstOPTICSOrderingPoints1999}
\bibinfo{author}{M.~Ankerst}, \bibinfo{author}{M.~M. Breunig}, \bibinfo{author}{H.-P. Kriegel}, \bibinfo{author}{J.~Sander},
\newblock \bibinfo{title}{{{OPTICS}}: Ordering points to identify the clustering structure},
\newblock \bibinfo{journal}{SIGMOD Rec.} \bibinfo{volume}{28} (\bibinfo{year}{1999}) \bibinfo{pages}{49--60}. \DOIprefix\doi{10.1145/304181.304187}.
\bibitem[{Ester et~al.(1996)Ester, Kriegel, Sander, and Xu}]{esterDensitybasedAlgorithmDiscovering1996}
\bibinfo{author}{M.~Ester}, \bibinfo{author}{H.-P. Kriegel}, \bibinfo{author}{J.~Sander}, \bibinfo{author}{X.~Xu},
\newblock \bibinfo{title}{A density-based algorithm for discovering clusters in large spatial databases with noise},
\newblock in: \bibinfo{booktitle}{Proceedings of the Second International Conference on Knowledge Discovery and Data Mining}, {{KDD}}'96, \bibinfo{publisher}{AAAI Press}, \bibinfo{address}{Portland, Oregon}, \bibinfo{year}{1996}, pp. \bibinfo{pages}{226--231}.
\bibitem[{Schubert et~al.(2017)Schubert, Sander, Ester, Kriegel, and Xu}]{schubertDBSCANRevisitedRevisited2017}
\bibinfo{author}{E.~Schubert}, \bibinfo{author}{J.~Sander}, \bibinfo{author}{M.~Ester}, \bibinfo{author}{H.~P. Kriegel}, \bibinfo{author}{X.~Xu},
\newblock \bibinfo{title}{{{DBSCAN Revisited}}, {{Revisited}}: {{Why}} and {{How You Should}} ({{Still}}) {{Use DBSCAN}}},
\newblock \bibinfo{journal}{ACM Trans. Database Syst.} \bibinfo{volume}{42} (\bibinfo{year}{2017}) \bibinfo{pages}{19:1--19:21}. \DOIprefix\doi{10.1145/3068335}.
\bibitem[{Murtagh and Contreras(2012)}]{murtaghAlgorithmsHierarchicalClustering2012}
\bibinfo{author}{F.~Murtagh}, \bibinfo{author}{P.~Contreras},
\newblock \bibinfo{title}{Algorithms for hierarchical clustering: An overview},
\newblock \bibinfo{journal}{WIREs Data Mining and Knowledge Discovery} \bibinfo{volume}{2} (\bibinfo{year}{2012}) \bibinfo{pages}{86--97}. \DOIprefix\doi{10.1002/widm.53}.
\bibitem[{{Vijaya} et~al.(2019){Vijaya}, Sharma, and Batra}]{vijayaComparativeStudySingle2019}
\bibinfo{author}{{Vijaya}}, \bibinfo{author}{S.~Sharma}, \bibinfo{author}{N.~Batra},
\newblock \bibinfo{title}{Comparative {{Study}} of {{Single Linkage}}, {{Complete Linkage}}, and {{Ward Method}} of {{Agglomerative Clustering}}},
\newblock in: \bibinfo{booktitle}{2019 {{International Conference}} on {{Machine Learning}}, {{Big Data}}, {{Cloud}} and {{Parallel Computing}} ({{COMITCon}})}, \bibinfo{publisher}{IEEE}, \bibinfo{address}{Faridabad, India}, \bibinfo{year}{2019}, pp. \bibinfo{pages}{568--573}. \DOIprefix\doi{10.1109/COMITCon.2019.8862232}.
\bibitem[{Fraley and Raftery(2002)}]{fraleyModelBasedClusteringDiscriminant2002}
\bibinfo{author}{C.~Fraley}, \bibinfo{author}{A.~E. Raftery},
\newblock \bibinfo{title}{Model-{{Based Clustering}}, {{Discriminant Analysis}}, and {{Density Estimation}}},
\newblock \bibinfo{journal}{Journal of the American Statistical Association} \bibinfo{volume}{97} (\bibinfo{year}{2002}) \bibinfo{pages}{611--631}. \DOIprefix\doi{10.1198/016214502760047131}.
\bibitem[{Jia et~al.(2014)Jia, Ding, Xu, and Nie}]{jiaLatestResearchProgress2014}
\bibinfo{author}{H.~Jia}, \bibinfo{author}{S.~Ding}, \bibinfo{author}{X.~Xu}, \bibinfo{author}{R.~Nie},
\newblock \bibinfo{title}{The latest research progress on spectral clustering},
\newblock \bibinfo{journal}{Neural Comput \& Applic} \bibinfo{volume}{24} (\bibinfo{year}{2014}) \bibinfo{pages}{1477--1486}. \DOIprefix\doi{10.1007/s00521-013-1439-2}.
\bibitem[{Ward(1963)}]{wardHierarchicalGroupingOptimize1963}
\bibinfo{author}{J.~H. Ward},
\newblock \bibinfo{title}{Hierarchical {{Grouping}} to {{Optimize}} an {{Objective Function}}},
\newblock \bibinfo{journal}{Journal of the American Statistical Association} \bibinfo{volume}{58} (\bibinfo{year}{1963}) \bibinfo{pages}{236--244}. \DOIprefix\doi{10.1080/01621459.1963.10500845}.
\bibitem[{James et~al.(2021)James, Witten, Hastie, and Tibshirani}]{jamesStatisticalLearning2021}
\bibinfo{author}{G.~James}, \bibinfo{author}{D.~Witten}, \bibinfo{author}{T.~Hastie}, \bibinfo{author}{R.~Tibshirani}, \bibinfo{title}{Statistical {{Learning}}}, \bibinfo{publisher}{Springer US}, \bibinfo{address}{New York, NY}, \bibinfo{year}{2021}, pp. \bibinfo{pages}{15--57}. \DOIprefix\doi{10.1007/978-1-0716-1418-1_2}.
\bibitem[{Tibshirani(1996)}]{tibshiraniRegressionShrinkageSelection1996}
\bibinfo{author}{R.~Tibshirani},
\newblock \bibinfo{title}{Regression shrinkage and selection via the lasso},
\newblock \bibinfo{journal}{Journal of the Royal Statistical Society. Series B (Methodological)} \bibinfo{volume}{58} (\bibinfo{year}{1996}) \bibinfo{pages}{267--288}. \DOIprefix\doi{10.1111/j.2517-6161.1996.tb02080.x}. \href{http://arxiv.org/abs/2346178}{{\tt arXiv:2346178}}.
\bibitem[{Hoerl and Kennard(1970)}]{hoerlRidgeRegressionBiased1970}
\bibinfo{author}{A.~E. Hoerl}, \bibinfo{author}{R.~W. Kennard},
\newblock \bibinfo{title}{Ridge {{Regression}}: {{Biased Estimation}} for {{Nonorthogonal Problems}}},
\newblock \bibinfo{journal}{Technometrics} \bibinfo{volume}{12} (\bibinfo{year}{1970}) \bibinfo{pages}{55--67}. \DOIprefix\doi{10.1080/00401706.1970.10488634}.
\bibitem[{Cortes and Vapnik(1995)}]{cortesSupportvectorNetworks1995}
\bibinfo{author}{C.~Cortes}, \bibinfo{author}{V.~Vapnik},
\newblock \bibinfo{title}{Support-vector networks},
\newblock \bibinfo{journal}{Mach Learn} \bibinfo{volume}{20} (\bibinfo{year}{1995}) \bibinfo{pages}{273--297}. \DOIprefix\doi{10.1007/BF00994018}.
\bibitem[{Breiman et~al.(1984)Breiman, Friedman, Olshen, and Stone}]{breimanClassificationRegressionTrees1984}
\bibinfo{author}{L.~Breiman}, \bibinfo{author}{J.~Friedman}, \bibinfo{author}{R.~A. Olshen}, \bibinfo{author}{C.~J. Stone}, \bibinfo{title}{Classification and {{Regression Trees}}}, \bibinfo{publisher}{{Chapman and Hall/CRC}}, \bibinfo{address}{New York}, \bibinfo{year}{1984}. \DOIprefix\doi{10.1201/9781315139470}.
\bibitem[{Breiman(2001)}]{breimanRandomForests2001}
\bibinfo{author}{L.~Breiman},
\newblock \bibinfo{title}{Random {{Forests}}},
\newblock \bibinfo{journal}{Machine Learning} \bibinfo{volume}{45} (\bibinfo{year}{2001}) \bibinfo{pages}{5--32}. \DOIprefix\doi{10.1023/A:1010933404324}.
\bibitem[{Friedman(2001)}]{friedmanGreedyFunctionApproximation2001}
\bibinfo{author}{J.~H. Friedman},
\newblock \bibinfo{title}{Greedy {{Function Approximation}}: {{A Gradient Boosting Machine}}},
\newblock \bibinfo{journal}{The Annals of Statistics} \bibinfo{volume}{29} (\bibinfo{year}{2001}) \bibinfo{pages}{1189--1232}. \href{http://arxiv.org/abs/2699986}{{\tt arXiv:2699986}}.
\bibitem[{Geurts et~al.(2006)Geurts, Ernst, and Wehenkel}]{geurtsExtremelyRandomizedTrees2006}
\bibinfo{author}{P.~Geurts}, \bibinfo{author}{D.~Ernst}, \bibinfo{author}{L.~Wehenkel},
\newblock \bibinfo{title}{Extremely randomized trees},
\newblock \bibinfo{journal}{Mach Learn} \bibinfo{volume}{63} (\bibinfo{year}{2006}) \bibinfo{pages}{3--42}. \DOIprefix\doi{10.1007/s10994-006-6226-1}.
\bibitem[{Chen and Guestrin(2016)}]{chenXGBoostScalableTree2016}
\bibinfo{author}{T.~Chen}, \bibinfo{author}{C.~Guestrin},
\newblock \bibinfo{title}{{{XGBoost}}: {{A Scalable Tree Boosting System}}},
\newblock in: \bibinfo{booktitle}{Proceedings of the 22nd {{ACM SIGKDD International Conference}} on {{Knowledge Discovery}} and {{Data Mining}}}, \bibinfo{publisher}{ACM}, \bibinfo{address}{San Francisco California USA}, \bibinfo{year}{2016}, pp. \bibinfo{pages}{785--794}. \DOIprefix\doi{10.1145/2939672.2939785}.
\bibitem[{Hastie et~al.(2009)Hastie, Tibshirani, and Friedman}]{hastieEnsembleLearning2009}
\bibinfo{author}{T.~Hastie}, \bibinfo{author}{R.~Tibshirani}, \bibinfo{author}{J.~Friedman},
\newblock \bibinfo{title}{Ensemble {{Learning}}},
\newblock in: \bibinfo{editor}{T.~Hastie}, \bibinfo{editor}{R.~Tibshirani}, \bibinfo{editor}{J.~Friedman} (Eds.), \bibinfo{booktitle}{The {{Elements}} of {{Statistical Learning}}: {{Data Mining}}, {{Inference}}, and {{Prediction}}}, \bibinfo{publisher}{Springer}, \bibinfo{address}{New York, NY}, \bibinfo{year}{2009}, pp. \bibinfo{pages}{605--624}. \DOIprefix\doi{10.1007/978-0-387-84858-7_16}.
\bibitem[{Aggarwal(2018)}]{aggarwalNeuralNetworksDeep2018}
\bibinfo{author}{C.~C. Aggarwal}, \bibinfo{title}{Neural {{Networks}} and {{Deep Learning}}: {{A Textbook}}}, \bibinfo{publisher}{Springer International Publishing}, \bibinfo{address}{Cham}, \bibinfo{year}{2018}. \DOIprefix\doi{10.1007/978-3-319-94463-0}.
\bibitem[{Lundberg et~al.(2020)Lundberg, Erion, Chen, DeGrave, Prutkin, Nair, Katz, Himmelfarb, Bansal, and Lee}]{lundbergLocalExplanationsGlobal2020}
\bibinfo{author}{S.~M. Lundberg}, \bibinfo{author}{G.~Erion}, \bibinfo{author}{H.~Chen}, \bibinfo{author}{A.~DeGrave}, \bibinfo{author}{J.~M. Prutkin}, \bibinfo{author}{B.~Nair}, \bibinfo{author}{R.~Katz}, \bibinfo{author}{J.~Himmelfarb}, \bibinfo{author}{N.~Bansal}, \bibinfo{author}{S.-I. Lee},
\newblock \bibinfo{title}{From local explanations to global understanding with explainable {{AI}} for trees},
\newblock \bibinfo{journal}{Nat Mach Intell} \bibinfo{volume}{2} (\bibinfo{year}{2020}) \bibinfo{pages}{56--67}. \DOIprefix\doi{10.1038/s42256-019-0138-9}.
\bibitem[{Zhang and Zhou(2014)}]{zhangReviewMultiLabelLearning2014}
\bibinfo{author}{M.-L. Zhang}, \bibinfo{author}{Z.-H. Zhou},
\newblock \bibinfo{title}{A {{Review}} on {{Multi-Label Learning Algorithms}}},
\newblock \bibinfo{journal}{IEEE Trans. Knowl. Data Eng.} \bibinfo{volume}{26} (\bibinfo{year}{2014}) \bibinfo{pages}{1819--1837}. \DOIprefix\doi{10.1109/TKDE.2013.39}.

\end{thebibliography}





\end{document}